%% file: neurips_2024.tex
\documentclass{article}

\input{support_files/setup_script}

\usepackage[preprint]{neurips_2024}

\usepackage[utf8]{inputenc} % allow utf-8 input
\usepackage[T1]{fontenc}  
\usepackage{floatrow} % use 8-bit T1 fonts
\usepackage{hyperref}       % hyperlinks
\usepackage{url}            % simple URL typesetting
\usepackage{booktabs}       % professional-quality tables
\usepackage{amsfonts}       % blackboard math symbols
\usepackage{nicefrac}       % compact symbols for 1/2, etc.
\usepackage{microtype}      % microtypography
\usepackage{xcolor}         % colors
\usepackage{wrapfig}
\usepackage{algorithmic}
\usepackage{algorithm}
\usepackage{lipsum}
\usepackage{array}
\usepackage{tcolorbox}

\title{Primal Dual Continual Learning: Balancing Stability and Plasticity through Adaptive Memory Allocation}

\author{Juan Elenter   \\
University of Pennsylvania \\
\texttt{elenter@seas.upenn.edu} 
\And
Navid NaderiAlizadeh \\
Duke University \\
\texttt{nnaderi@seas.upenn.edu}
\And
Tara Javidi \\
University of California, San Diego\\
\texttt{tjavidi@ucsd.edu}
\And
Alejandro Ribeiro\\
University of Pennsylvania\\
\texttt{aribeiro@seas.upenn.edu}
}

\begin{document}
\maketitle

\begin{abstract}
Continual learning is inherently a constrained learning problem. The goal is to learn a predictor under a \emph{no-forgetting} requirement. Although several prior studies formulate it as such, they do not solve the constrained problem explicitly. In this work, we show that it is both possible and beneficial to undertake the constrained optimization problem directly. To do this, we leverage recent results in constrained learning through Lagrangian duality. We focus on memory-based methods, where a small subset of samples from previous tasks can be stored in a replay buffer. In this setting, we analyze two versions of the continual learning problem: a coarse approach with constraints at the task level and a fine approach with constraints at the sample level. We show that dual variables indicate the sensitivity of the optimal value of the continual learning problem with respect to constraint perturbations. We then leverage this result to partition the buffer in the coarse approach, allocating more resources to harder tasks, and to populate the buffer in the fine approach, including only impactful samples. We derive a deviation bound on dual variables as sensitivity indicators, and empirically corroborate this result in diverse continual learning benchmarks. We also discuss the limitations of these methods with respect to the amount of memory available and the expressiveness of the parametrization.
\end{abstract}

\input{body}

\bibliography{references}
\bibliographystyle{support_files/bib_style}

\appendix
\include{appendix}

%%%%%%%%%%%%%%%%%%%%%%%%%%%%%%%%%%%%%%%%%%%%%%%%%%%%%%%%%%%%

%\input{checklist}

\end{document}

%% file: support_files/setup_script.tex
\input{support_files/conference_import_replacement}

\input{support_files/our_math_commands}

\definecolor{linkcolor}{RGB}{0,120,130}

\usepackage[colorlinks=true,
    linkcolor=linkcolor,
    citecolor=linkcolor,
    filecolor=linkcolor,
    urlcolor=linkcolor,
    pagebackref=true]{hyperref} 
\usepackage{url}

\renewcommand*\backref[1]{\ifx#1\relax \else (Cit. on p. #1) \fi}

\usepackage[capitalize]{cleveref}

\Crefname{algorithm}{Algo.}{Algos.}
\Crefname{theorem}{Thm.}{Thms.}
\Crefname{lemma}{Lem.}{Lems.}
\Crefname{proposition}{Prop.}{Props.}
\Crefname{appendix}{Appx.}{Appx.}

\theoremstyle{plain}
\newtheorem{theorem}{Theorem}

\newtheorem{proposition}{Proposition}
\newtheorem{lemma}{Lemma}
\newtheorem{corollary}{Corollary}
\theoremstyle{definition}
\newtheorem{definition}{Definition}
\newtheorem{assumption}{Assumption}

\theoremstyle{remark}

\theoremstyle{problem}

\usepackage[textsize=tiny]{todonotes}
\usepackage{mathtools}
\usepackage{multirow}
\usepackage{booktabs}
\usepackage{adjustbox}

\usepackage{caption}
\usepackage{subcaption}

\usepackage{tikz}
\usepackage{pgfplots}
\pgfplotsset{compat=1.17}
\usepackage{standalone}
\usetikzlibrary{shapes,arrows,fit,positioning, calc, decorations.markings}
\definecolor{captiongray}{RGB}{100,100,100}


\usepackage{minitoc}

%% file: support_files/conference_import_replacement.tex
\usepackage{natbib}
\setcitestyle{authoryear,round,citesep={;},aysep={,},yysep={;}}

\usepackage[dvipsnames]{xcolor}

%% file: support_files/our_math_commands.tex
\usepackage{amsfonts}
\usepackage{nicefrac}       % compact symbols for 1/2, etc.
\usepackage{amsmath}
\usepackage{amsthm}
\usepackage{amssymb}
\usepackage{cancel}
\usepackage{mathtools}
\usepackage{bbm}
\usepackage{bm}

\usepackage{cases}         % Numbered cases

\usepackage{mathalpha}

\usepackage{microtype}      % microtypography
\usepackage{enumitem}

\usepackage{pifont}

\newcommand{\vepsilon}{\boldsymbol{\epsilon}}

\newcommand{\algcomment}[1]{\hfill// {\small\texttt{#1}}}

\definecolor{lightgray}{RGB}{230, 230, 230}
\definecolor{mathred}{RGB}{204, 69, 90}
\definecolor{mathblue}{RGB}{4, 78, 112}
\definecolor{mathgreen}{RGB}{1, 135, 70}

\DeclareMathOperator*{\argmax}{arg\,max}
\DeclareMathOperator*{\argmin}{arg\,min}

%% file: body.tex
\setlength{\belowdisplayskip}{3pt}

\section{Continual Learning is a Constrained Learning Problem}

%\lipsum[1]

In real-world settings, agents must adapt to a dynamic stream of observations they receive from their environment. This necessity has led to extensive research in \emph{continual learning}, where the goal is to train agents to sequentially solve a diverse set of tasks~\citep{Thrun-1995-16134}.

Given its sequential nature and the finite capacity of machine learning (ML) models, the primary challenge in continual learning lies in balancing the acquisition of new knowledge (plasticity) with the retention of previously integrated knowledge (stability). Mishandling the stability-plasticity balance can result in significant performance degradation on prior tasks.  Avoiding this phenomenon, termed \emph{catastrophic forgetting}, naturally leads to constrained optimization formulations, which have appeared extensively in the continual learning literature~\citep{gss, agem, gem, peng2023ideal}. 

Most approaches do not solve this constrained optimization problem explicitly. Instead, they use gradient projections~\citep{gem, agem}, promote proximity in the parameter space~\citep{proxim, ewc}, or penalize deviations from a reference model. For instance, when fine-tuning a large language model (LLM) using RLHF, deviations from the pre-trained model are penalized to avoid degradation in text completion \citep{christiano2017deep}. This work demonstrates that it is both possible and beneficial to undertake the constrained learning problem directly (\textbf{Contribution 1}). To do this, we leverage recent advances in constrained learning through Lagrangian duality \citep{ConstrainedNonConvex} and build a framework that contemplates both task-level and instance-level forgetting.

State-of-the-art continual learning methods often include replay buffers, in which agents store a small subset of the previously seen instances. These methods have become ubiquitous, as they generally outperform their memoryless counterparts \citep{masana2022class, survey1, de2021continual}. The \emph{principled} constrained learning framework proposed in this paper enables an adaptive and efficient management of the memory buffer. In this framework, the continual learning problem is viewed through the lens of constrained learning via Lagrangian duality. This perspective provides access to the sensitivity of the optimal value of the continual learning problem with respect to constraint perturbations, indicating how performance on the current task is influenced by the difficulty of not forgetting past tasks. At the task level, we leverage this result to partition the buffer, allocating more resources to harder tasks and using dual variables as adaptive regularization weights for the replay losses (\textbf{Contribution 2}). At the sample level, we use this sensitivity information to populate the buffer, including only impactful instances. These techniques provide a direct handle on the stability-plasticity trade-off incurred in the learning process (\textbf{Contribution 3}).

A continual learner aims to minimize the expected risk over a set of \emph{tasks}, \vspace{-0.07in} \begin{equation*} f^{\star}_{\theta} = \argmin_{\theta \in \Theta} \sum^{T}_{t=1} \mathbb{E}_{\mathfrak{D}_t} \left[  \ell(f_{\theta}(x), y) \right],
\end{equation*} where $T$ is the number of tasks, $\mathfrak{D}_t$ is the data distribution associated to task $t$ and $f_{\theta}:\mathcal{X} \to \mathcal{Y}$ is the function associated with parameters $\theta \in \Theta \subseteq \mathbb{R}^p$. The tasks and their corresponding data distributions are observed sequentially. That is, at time $t$, data from previous tasks (i.e., $\mathfrak{D}_1, \cdots, \mathfrak{D}_{t-1}$) and from future tasks (i.e., $\mathfrak{D}_{t+1}, \cdots, \mathfrak{D}_{T}$) are not available. In this setting, the main issue that arises is catastrophic forgetting: if we sequentially fine-tune $f$ on each incoming distribution, the performance on previous tasks could drop severely. A continual learner is one that is stable enough to retain acquired knowledge and malleable enough to gain new knowledge. 

If the no-forgetting requirement is enforced at the task level, we can formulate the continual learning problem as minimizing the statistical risk on the current task without harming the performance of the model on previous tasks, i.e., \begin{align*} \vspace{-0.05in}
\label{Pt}
\tag{$P_t$}
P_t = & \min_{\theta \in \Theta} \:\:\: \mathbb{E}_{\mathfrak{D}_t} [ \ell(f_\theta(x), y)  ], \\
& \: \text{s.t. } \quad \mathbb{E}_{\mathfrak{D}_k} [ \ell(f_\theta(x), y) ] \leq \epsilon_k, \quad \forall \: k\in\{1,\dots,t-1\},
\end{align*}
where $\epsilon_k \in \mathbb{R}$ is the \emph{forgetting tolerance} of task $k$, i.e., the worst average loss that is admissible in a certain task. In many cases, this is a \emph{design} requirement, and not a tunable parameter. For instance, in medical applications, $\epsilon_k$ can be tied to regulatory constraints. If the upper bound is set to the unconstrained minimum (i.e., $\epsilon_k = \min_{\theta \in \Theta} \mathbb{E}_{\mathfrak{D}_k} [ \ell(f_{\theta}(x), y) ]$), then the solution to problem (\ref{Pt}) corresponds to an \emph{ideal continual learner} \citep{peng2023ideal}. However, we do not have access to $\mathfrak{D}_k$ for $k \neq t$, but only to a \emph{memory buffer} $\mathcal{B}_t = \cup_{k=1}^{t-1} \mathcal{B}_k(t)$, where $\mathcal{B}_k(t)$ denotes the subset of the buffer allocated to task $k$ while observing task $t$. When possible, we will obviate the dependence on the index $t$ to ease the notation.

In this setting, the main questions that arise are: (i) When is the constrained learning problem~\eqref{Pt} solvable? (ii) How to solve it? (iii) How to partition the buffer $\mathcal{B}$ across the different tasks ? (iv) Which samples from each task should be stored in the buffer?

This paper is structured as follows: in Section \ref{sec:cl}, we present the duality framework used to undertake the constrained learning problem. In Section \ref{sec:pdcl}, we characterize the variations of the optimal value of the continual learning problem, which leads to the proposed buffer partition strategy. In Section \ref{sec:sample_selec}, we discuss sample selection leveraging the information carried by dual variables; and in Section \ref{sec:emp_valid} we present numerical results in image, audio and medical datasets that validate these findings.

\section{Continual Learning in the Dual Domain}
\label{sec:cl}

For continual learning to be justified, tasks need to be similar. The following assumption characterizes this similarity in terms of the distance between the set of optimal predictors associated to each task.

\begin{assumption}\label{ass:task_sim} Let $\mathcal{F}_t^{\star} = \{ \theta \in \Theta : \mathbb{E}_{\mathfrak{D}_t} [ \ell(f_{\theta}(x), y) ] = \min_{\theta \in \Theta}\mathbb{E}_{\mathfrak{D}_t} [ \ell(f_{\theta}(x), y) ] \} $ be the set of optimal predictors associated to task $t$. The pairwise distance between optimal sets is bounded as in $$ d(\mathcal{F}^{\star}_i, \mathcal{F}^{\star}_j) \leq \delta, \quad \forall i,j \in \{1,\cdots,T \}.$$
\end{assumption}
Several task similarity assumptions have been proposed in the literature, most of which can be formulated as Assumption \ref{ass:task_sim} with an appropriate choice of $d(\cdot ,\cdot )$ and $\delta$. In this work, we use the standard (Haussdorf) distance between non-empty sets: $d(X, Y)=\max \left\{\sup _{x \in X} d(x, Y), \sup _{y \in Y} d(X, y)\right\}$. Note that in over-parameterized settings, deep neural networks attain near-interpolation regimes and this assumption is not strict \citep{liu2021loss}. This leads to the following proposition, which characterizes the feasibility of the continual learning problem. In particular, it suggests that for Problem \eqref{Pt} to be feasible, the forgetting tolerances $\{\epsilon_k\}_{k=1}^T$ need to match the task similarity $\delta$.
\begin{proposition} 
\label{prop:set_eps}
Let $m_k$ be the unconstrained minimum associated to task $k$ and let $M$ be the Lipschitz constant of the loss $\ell(\cdot, y)$. Under Assumption~\ref{ass:task_sim}, there exists $\theta \in \Theta$ such that, 
\begin{equation*}
\mathbb{E}_{\mathfrak{D}_k} [ \ell(f_{\theta}(x), y) ] \leq m_k + \frac{T-1}{T}M\delta, \quad \forall k\in\{1,\cdots,T\}.
\end{equation*}
\end{proposition}
For instance, if $\epsilon_k = m_k+M\delta$ for all $k$, then problem \eqref{Pt} is feasible at all iterations, and its solution is $M\delta$ close to the optimum in terms of the expected loss on the current task. 

However, the feasibility of problem \eqref{Pt} tells us nothing about how to solve it. Note that even if the loss $\ell(f_{\theta}(x), y)$ is convex in~$(f_{\theta}(x), y)$~(such as MSE and cross-entropy loss), the function~$\ell$ need not be convex in~$\theta$. This is the case, for instance, for typical modern ML models~(e.g., if $f_{\theta}$ is a convolutional neural network or a transformer-based language model). Hence, \eqref{Pt} is usually a non-convex optimization problem for which there is no straightforward way to project onto the feasibility set~(i.e., onto the set of no-forgetting models).

In light of these challenges, we turn to Lagrangian duality. \eqref{Pt} is a statistical constrained optimization problem, whose empirical dual can be written as 
\begin{equation}
\tag{$D_t$}\label{Dt}
D_t = \max_{\boldsymbol{\lambda} \in \mathbb{R}_+^{t-1}} \min_{\theta \in \theta} \hat{\mathcal{L}}(\theta, \boldsymbol{\lambda}) := \frac{1}{n_t}\sum_{i=1}^{n_t} [ \ell(f_{\theta}(x_i), y_i) ] + \sum_{k=1}^{t-1} \lambda_k \left( \frac{1}{n_k} \sum_{i=1}^{n_k} [ \ell(f_{\theta}(x_i), y_i) ] - \epsilon_k \right),    
\end{equation} where $\hat{\mathcal{L}}(\theta, \boldsymbol{\lambda})$ denotes the empirical Lagrangian, $n_k$ denotes the number of samples from task $k$ available at iteration $t$, and  $\boldsymbol{\lambda}$ %$=[\lambda_1 ~ \dots ~ \lambda_t]^T$
denotes the vector of dual variables corresponding to the task-level constraints. For a fixed $\boldsymbol{\lambda}$, the Lagrangian $\hat{\mathcal{L}}(\theta, \boldsymbol{\lambda})$ is a regularized objective, where the losses on previous tasks act as regularizing functionals. Thus, the saddle point problem in \eqref{Dt} can be viewed as a two-player game, or as a regularized minimization, where the regularization weight $\boldsymbol{\lambda}$ is updated during the training procedure according to the degree of constraint satisfaction or violation. We elaborate on this iterative procedure when we present the primal-dual algorithm in Section \ref{sec:pdcl}.

Note that \emph{the weighing of the replayed losses relative to the loss on the current task is determined by the dual variables $\boldsymbol{\lambda}$}, rather than by a hyperparameter or by the number of tasks seen so far \citep{buzzega2020dark, MICHIELI2021103167}. Albeit simple, this is a key difference concerning previous replay and knowledge distillation approaches. In the sequel, we show that optimal dual variables $\boldsymbol{\lambda}^{\star}$ indicate the sensitivity of the optimal value $P_t$ as a function of the constraint levels $\{\epsilon_k\}_{k=1}^{t-1}$ and can thus be used as indicators of task difficulty.
\begin{algorithm}[t]
\caption{Primal-Dual Continual Learning (PDCL)}
    \label{alg:pdcl}
    \begin{algorithmic}[1]
    \STATE {\bfseries Input:} Num. Tasks $T$, primal (dual) learning rate $\eta_p$ ($\eta_d$), Number of primal steps 
 per dual step $T_p$ , constraint levels $\{\epsilon_k\}_{k=1}^T$, number of iterations $n_{\text{iter}}$ .
    \STATE Initialize $\theta$
    \FOR{$t=1, \ldots, T$}
    \STATE Initialize $\boldsymbol{\lambda}$
     \FOR{$i=1, \ldots, n_{\text{iter}}$}
        \STATE 
        $\theta \: \leftarrow \: \theta - \eta_p \nabla_{\theta}\mathcal{L}(\theta, \boldsymbol{\lambda}) \quad \quad (\times \: T_p)
        $ \algcomment{Update primal variables}
        \STATE
        $
        s_{k} \leftarrow \frac{1}{n_k} \sum_{j=1}^{n_k} \ell(f_{\theta}(x_j), y_j) - \epsilon_k, \forall k\in\{1,\dots,t-1\}
        $ \algcomment{Evaluate constraint slacks}
        \STATE
        $
        \lambda_{k} \leftarrow \left[\lambda_{k} + \eta_d s_{k} \right]_{+}, \forall k\in\{1,\dots,t-1\}
        $ \algcomment{Update dual variables}
    \ENDFOR
    \STATE 
    $n^{\star}_1,\dots,n^{\star}_t \: \leftarrow \: PB(\lambda_1, \dots, \lambda_{t-1}) $ \algcomment{Partition Buffer}
    \STATE 
    $
    \mathcal{B}_t \: \leftarrow \: FB(\mathcal{B}_{t-1}, \mathfrak{D}_t, \{n^{\star}_k\}_{k=1}^t)
    $
    \algcomment{Fill Buffer}
    \ENDFOR
    \STATE {\bfseries Return:}{  $\theta$, $\boldsymbol{\lambda}$.} 
    \end{algorithmic}
\end{algorithm}

\section{Primal Dual Continual Learning}
\label{sec:pdcl}
Provided we have enough samples per task and the parameterization $\Theta$ is rich enough, \eqref{Dt} can approximate the constrained statistical problem \eqref{Pt}. More precisely, the empirical duality gap, defined as the difference between the optimal value $D_t$ of the empirical dual and the statistical primal $P_t$, is bounded \citep[Theorem 1]{ConstrainedNonConvex}. Furthermore, the dual function \begin{equation*} \hat{g}(\boldsymbol{\lambda}) = \min _{\theta \in \theta} \hat{\mathcal{L}}(\theta, \boldsymbol{\lambda}) \end{equation*} is the minimum of a family of affine functions on $\boldsymbol{\lambda}$, and thus is concave, irrespective of whether (\ref{Pt}) is convex. As such, though $\hat{g}$ may not be differentiable, it can be equipped with \emph{supergradients} that provide potential ascent directions. Explicitly, a vector $s \in \mathbb{R}^m$ is a supergradient of the concave function $\hat{g}$ at a point $\boldsymbol{\lambda}_1$ if $\hat{g}(\boldsymbol{\lambda}_2) - \hat{g}(\boldsymbol{\lambda}_1) \geq s^T(\boldsymbol{\lambda}_2-\boldsymbol{\lambda}_1)$ for all $\boldsymbol{\lambda}_2$. The set of all supergradients of $\hat{g}$ at $\boldsymbol{\lambda}_1$ is called the \emph{superdifferential} and is denoted $\partial \hat{g}(\boldsymbol{\lambda}_1)$. When the loss $\ell$ is continuous, the superdifferential of $\hat{g}$ admits a simple description \citep{shor2013nondifferentiable}, namely, \begin{equation*}  \partial \hat{g}(\boldsymbol{\lambda}) = \text{conv} \big[ L(f_{\theta}(\boldsymbol{\lambda})): f_{\theta}(\boldsymbol{\lambda}) \in \mathcal{F}_{\theta}^{\star}(\boldsymbol{\lambda}) \big] , \end{equation*} where $ L(f) := \left[ \frac{1}{n_k} \sum_{i=1}^{n_k} [ \ell(f_{\theta}(x_i), y_i) ] - \epsilon_k \right]_{k=1}^{t-1} $ , $\text{conv}(\mathcal{S})$ denotes the convex hull of the set $\mathcal{S}$ and $\mathcal{F}_{\theta}^{\star}(\boldsymbol{\lambda})=\argmin_{\theta} \hat{\mathcal{L}}(f_{\theta}, \boldsymbol{\lambda})$ is the set of Lagrangian minimizers $f_{\theta}(\boldsymbol{\lambda})$ associated to $\boldsymbol{\lambda}$.

Consequently, the outer problem (i.e., $\max_{\boldsymbol{\lambda}\succeq0} \hat{g}(\boldsymbol{\lambda})$) corresponds to the maximization of a concave function and can be solved via super-gradient ascent \citep{asumanprimalrates}. The inner minimization, however, is generally non-convex, but there is ample empirical evidence that deep neural networks can attain \textit{good} local minima when trained with stochastic gradient descent \citep{rethink_gen}. Hence, the saddle-point problem \eqref{Dt} can be undertaken by alternating the minimization with respect to $\theta$ (line 6 in Alg. \ref{alg:pdcl}) and the maximization with respect to $\boldsymbol{\lambda}$ (line 8 in Alg. \ref{alg:pdcl})  \citep{arrowhurwitz, shor2013nondifferentiable}. We refer to \citep{ConstrainedNonConvex} and \citep{elenter2024nearoptimal} for a thorough analysis on the convergence and primal recovery properties of this procedure in the context of constrained learning.

An overview of the proposed primal-dual continual learning method (PDCL) is provided in Algorithm~\ref{alg:pdcl}. Note that $PB$ (Buffer Partition) denotes a generic procedure to compute the number of samples allocated to each task at time $t$ given a vector of dual variables $\boldsymbol{\lambda}$. Similarly, $FB$ (Fill Buffer) denotes a generic mechanism for populating the buffer given a specific memory partition $\{n_1, \cdots, n_t \}$. Recall that at iteration $t$, the only samples available are the ones from the current task and those stored in the buffer $\mathcal{B}_t = \cup_{k=1}^{t-1} \mathcal{B}_k(t)$, with $n_k(t) = |\mathcal{B}_k(t)|$.

\subsection{Dual Variables Capture the Stability-Plasticity Trade-off}
\label{sec:duals_stabplast}
The Buffer Partition method $(PB)$ takes as input the vector of dual variables $\boldsymbol{\lambda}$ because they indicate relative task difficulty. In this section, we formalize this result using tools from convex variational analysis. Specifically, we leverage the fact that, though non-convex and intricate, the continual learning problem \eqref{Pt} is the parametrized version of a benign functional optimization problem.  

The \emph{unparametrized} constrained learning problem is defined as
\begin{align*}
\label{Ptilde}
\tag{$\tilde{P}_t$}
\tilde{P}_t = & \min_{\phi \in \mathcal{F}} \:\:\: \mathbb{E}_{\mathfrak{D}_t} [ \ell(\phi(x), y)  ], \\
& \, \text{s.t. } \quad \mathbb{E}_{\mathfrak{D}_k} [ \ell(\phi(x), y) ] \leq \epsilon_k, \quad \forall \: k\in\{1,\dots,t-1\},
\end{align*} where $\mathcal{F}$ denotes a compact functional space satisfying $\|\phi\|_{L_2} \leq$ $R$ for every $\phi \in \mathcal{F}$. For instance, $\mathcal{F}$ can be a subset of the space of continuous functions on a compact set or a reproducing kernel Hilbert space (RKHS) and $\mathcal{F}_{\theta}= \{ f_{\theta} : \theta \in \Theta \}$ can be induced by a neural network architecture with smooth activations or a finite linear combination of kernels. %In both cases, we know that $\mathcal{F}_{\theta}$ can uniformly approximate $\mathcal{F}$ arbitrarily well as the dimension of $\theta$ grows \citep{hornik1991approximation, berlinet2011reproducing}. 
The smallest choice of $\mathcal{F}$ is in fact $\overline{\text{conv}}(\mathcal{F}_{\theta})$ (closed convex hull of $\mathcal{F}_{\theta}$). Analogous to the definition from Section \ref{sec:cl},
\begin{equation}
\tag{$\tilde{D}_t$}
\label{Dtilde}
\tilde{D}_t = \max_{\boldsymbol{\tilde{\lambda}} \in \mathbb{R}_+^{t-1}} \min_{\phi \in \mathcal{F}} \mathcal{L}(\phi, \boldsymbol{\tilde{\lambda}}) := \mathbb{E}_{\mathfrak{D}_t} [ \ell(\phi(x), y) ] + \sum_{k=1}^{t-1} {\tilde{\lambda}}_{k} \left( \mathbb{E}_{\mathfrak{D}_k} [ \ell(\phi(x), y) ] - \epsilon_k \right),   
\end{equation} is the \emph{unparametrized} empirical dual problem. The only difference between problems (\ref{Pt}) and (\ref{Ptilde}) is the set over which the optimization is carried out. Thus, if the parametrization $\Theta$ is rich enough (e.g., deep neural networks such as large transformers), the set $\mathcal{F}_{\theta}$ is essentially the same as $\mathcal{F}$, and we should expect the properties of the solutions $\boldsymbol{\lambda}^{\star}$ and $\boldsymbol{\tilde{\lambda}}^{\star}$ to problems \eqref{Dt} and \eqref{Dtilde} to be similar. This insight leads us to the $\nu-$near universality of the parametrization assumption.

\begin{assumption}
\label{ass:nu}  For all $\phi \in \mathcal{F}$, there exists $\theta \in \Theta$ such that $\| \phi - f_{\theta} \|_{L_2} \leq \nu$.
\end{assumption}
\begin{wrapfigure}{r}{0.36\textwidth}
  \begin{center}
  \vspace{-0.15in}
    \includegraphics[width=\textwidth]{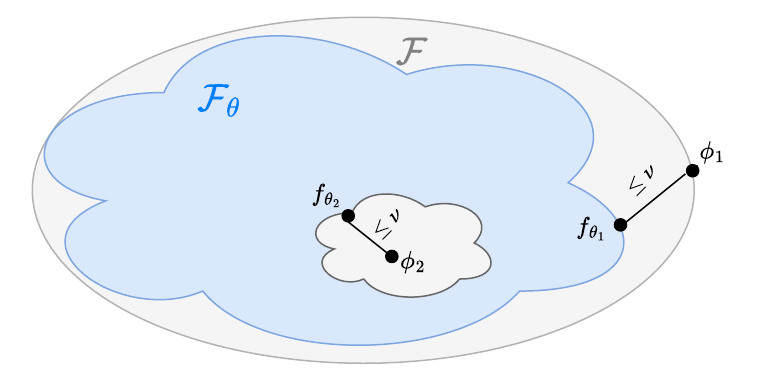}
  \end{center}
  \caption{Diagram of a $\nu-$Universal Parametrization $\mathcal{F}_{\Theta}$ of $\mathcal{F}$.}
  \vspace{-0.2in}
\end{wrapfigure}

The constant $\nu$ in Assumption \ref{ass:nu} is a measure of how well $\mathcal{F}_{\theta}$ covers $\mathcal{F}$. Consider, for instance, that $\mathcal{F}$ is the set of continuous functions on a compact set and $\mathcal{F}_{\theta}$ the set of functions implementable with a two-layer neural network with sigmoid activations and $K$ hidden neurons. If the parametrization has $10$ neurons in the hidden layer, it is considerably worse at representing elements in $\mathcal{F}$ than one with $1000$ neurons. While determining the exact value of $\nu$ is, in general, not straightforward, any $\nu>0$ can be achieved for a large enough number of neurons  \citep{hornik1991approximation}. The same holds for the number of kernels and an RKHS \citep{berlinet2011reproducing}.  

%\begin{figure}[t]
%   \centering
%    \adjustbox{trim=1mm 10mm 2mm 6mm, clip}{\includegraphics[width=0.36\textwidth]{figures/general/pertsubgrad.pdf}}
%    \adjustbox{trim=1mm 10mm 2mm 6mm, clip}{\includegraphics[width=0.36\textwidth]{figures/general/wpertsubgrad.pdf}}
%    \caption{Sensitivity information captured by $\boldsymbol{\lambda}^{\star}$ (Theorem \ref{theo:omegasubdiff})}
%\end{figure}

To formalize the fact that $\boldsymbol{\lambda}^{\star}$ measures relative task difficulty, we combine two results. First, optimal duals $\boldsymbol{\tilde{\lambda}}^{\star}$ of the \emph{unparametrized} problem \eqref{Dtilde} capture the sensitivity of $\tilde{P}_t$ with respect to constraint perturbations (see Appendix \ref{app:sensidual}) and second, dual variables $\boldsymbol{\lambda}^{\star}$ of the \emph{empirical parametrized} problem \eqref{Dt} can not be too far from $\boldsymbol{\tilde{\lambda}}^{\star}$ (see Appendix \ref{app:distduals}). We now state an assumption (specifically a constraint qualification) guaranteeing that the aforementioned results hold.

\begin{assumption}\label{ass:const_qual} The loss $\ell$ is $M-$Lipschitz and convex, the functional space $\mathcal{F}$ is convex, and there exists a strictly feasible solution (i.e., $\exists \: \phi\in\mathcal{F}$ such that $\mathbb{E}_{\mathfrak{D}_k} [ \ell(f(x), y) ] < \epsilon_k, \: \forall k$).
\end{assumption}
Note that we require convexity of the losses with respect to their functional arguments and not model parameters $\theta$, which holds for most typical losses, e.g., mean squared error and cross-entropy loss. Along with the boundedness of $\mathcal{F}$, this assumption guarantees uniform convergence \citep[Theorem 5]{shalev2009stochastic}, implying that with probability at least $1-\delta$, for all $\phi\in\mathcal{F}$ we have:  
\begin{equation}
    \left| \mathbb{E}_{\mathfrak{D}_k} [ \ell(f(x), y) ]-\frac{1}{n_k} \sum_{i=1}^{n_k}  \ell(f(x_i), y_i)  \right|  \leq \zeta\left(n_k, \delta\right), ~ \forall k=1, \dots, T, 
\end{equation} where the sample complexity function $\zeta\left(n_k, \delta \right)$ is of order $\mathcal{O}\left( R M \sqrt{d \log \left(n_k \right) \log (d / \delta)}^{} / {}\sqrt{n_k} \right) .$ Additionally, Assumption \ref{ass:const_qual} characterizes the curvature $\mu$ of the \emph{unparametrized} dual function $\tilde{g}(\boldsymbol{\lambda}) = \min_{\phi \in \mathcal{F}}\mathcal{L}(\phi, \boldsymbol{\tilde{\lambda}})$, on which we elaborate in Appendix \ref{app:curvdual}, and leads to the following theorem. %Analyzing the variations of the optimal value of the continual learning problem \eqref{Ptilde} a function $\tilde{P}_t(\mathbf{\epsilon})$ of the constraint upper bounds (or forgetting tolerances) $\mathbf{\epsilon}$ leads to the following result.

\begin{tcolorbox}[arc=0mm, colback=blue!5!white,colframe=blue!75!black]
\begin{theorem}
\label{theo:omegasubdiff}
Under Assumptions \ref{ass:nu} and \ref{ass:const_qual}, with probability at least $1-t\delta$, $\boldsymbol{\lambda}^{\star}$ belongs to the $\omega-$subdifferential of $\Tilde{P}_t(\epsilon)$ at $\epsilon$. That is:
\begin{equation*}
  - \boldsymbol{\lambda}^{\star} \: \in \: \partial_{\omega} \Tilde{P}_t(\epsilon),
\end{equation*} with the constant $\omega^2= \frac{2}{\mu}\left[ M\nu(1+\|\boldsymbol{\lambda}^{\star}\|_1) + 6\zeta(\Tilde{n}, \delta)(1+ \Delta) \right]$, the sensitivity parameter $ \Delta~=~\max \{ \| \boldsymbol{\tilde{\lambda}}^{\star} \|_1, \| \boldsymbol{\lambda}^{\star} \|_1 \}$, and the sample complexity given by $ \Tilde{n} = \min_{i=1,\dots,T} n_i(T)$.

\end{theorem}
\end{tcolorbox}
\begin{figure}
    \centering
    \adjustbox{trim=1mm 10mm 2mm 6mm, clip}{\includegraphics[width=0.28\linewidth]{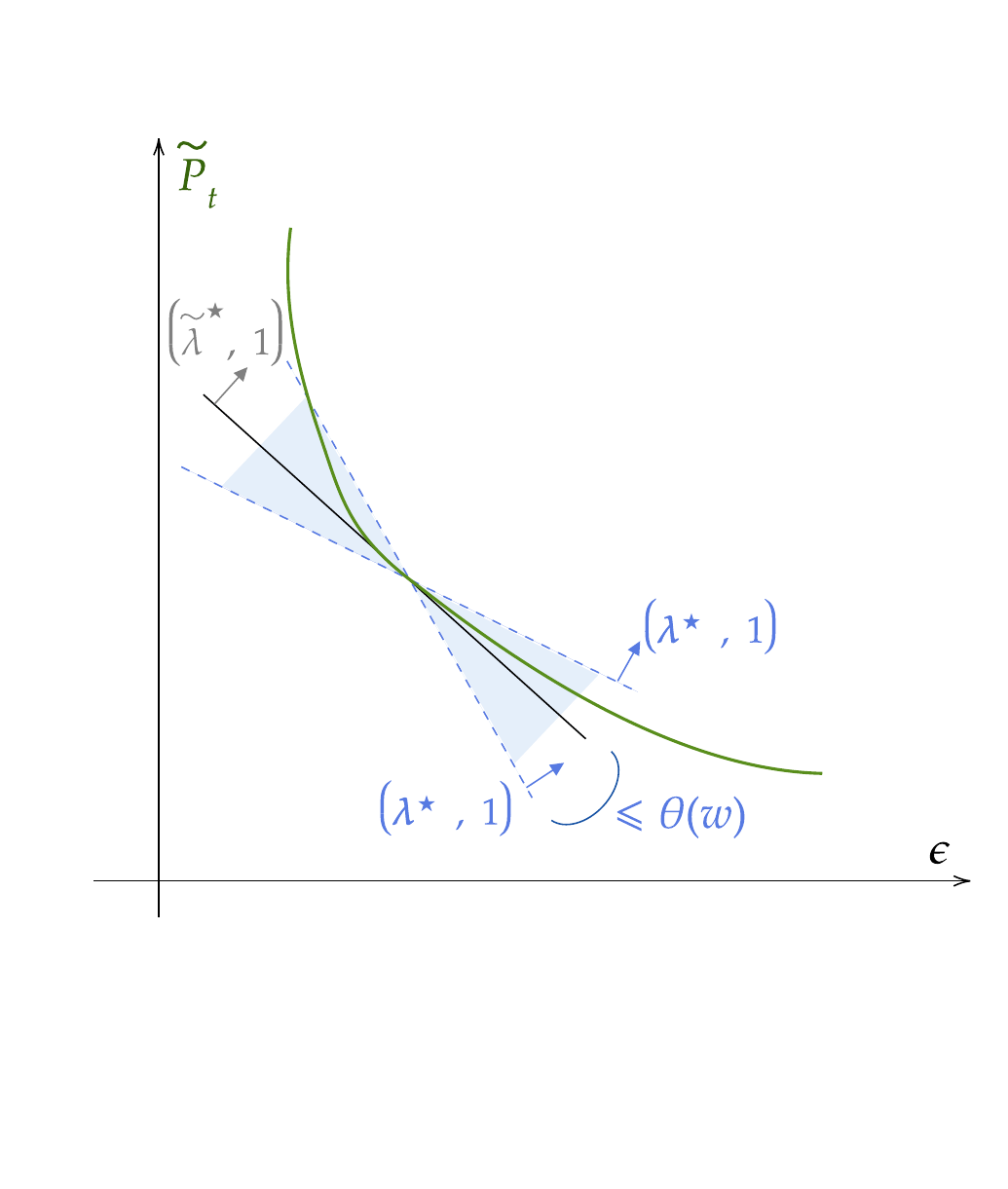}}
    \adjustbox{trim=1mm 10mm 2mm 6mm, clip}{\includegraphics[width=0.28\linewidth]{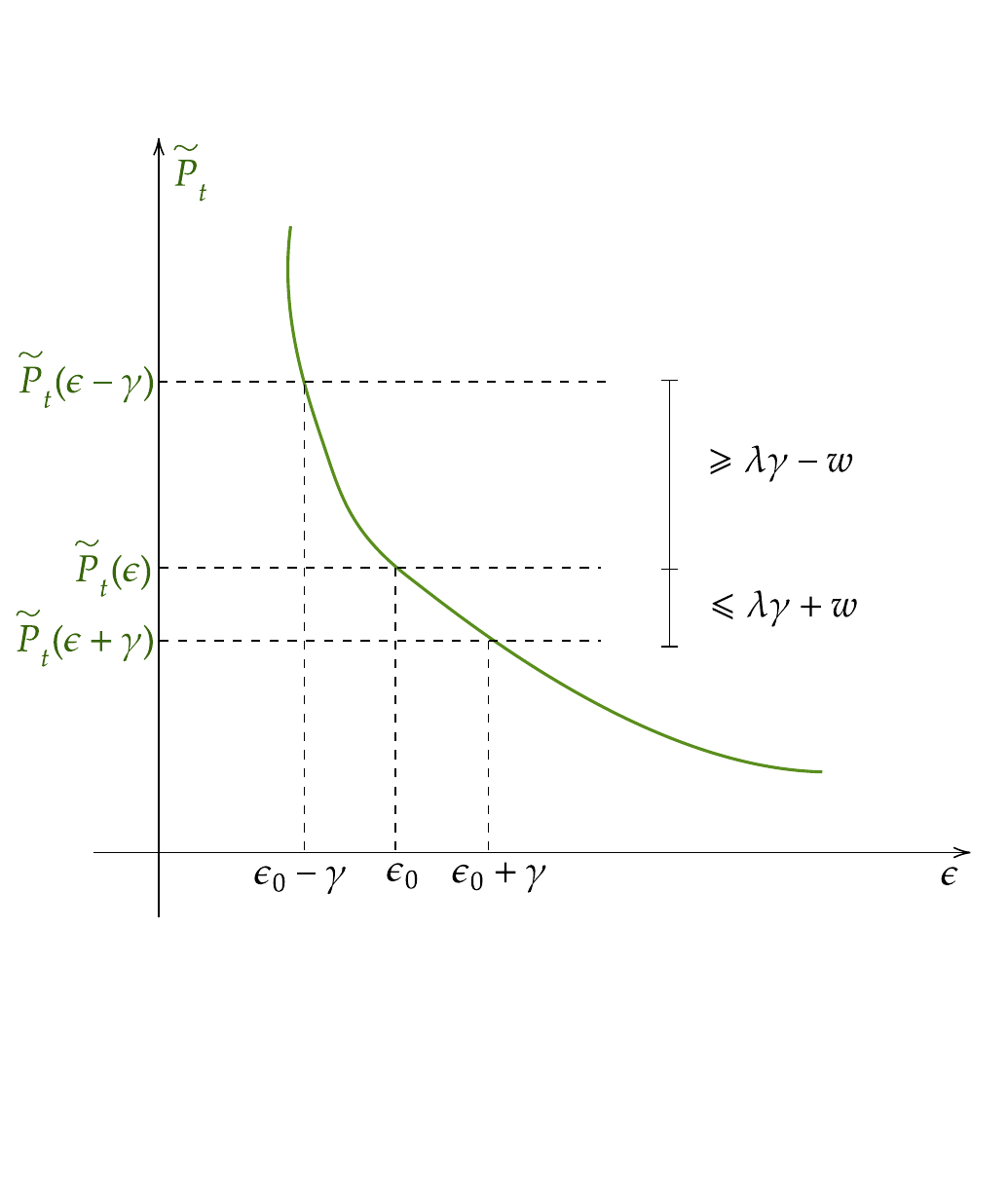}}
    \hspace{0.3in}
    \includegraphics[width=0.36\linewidth]{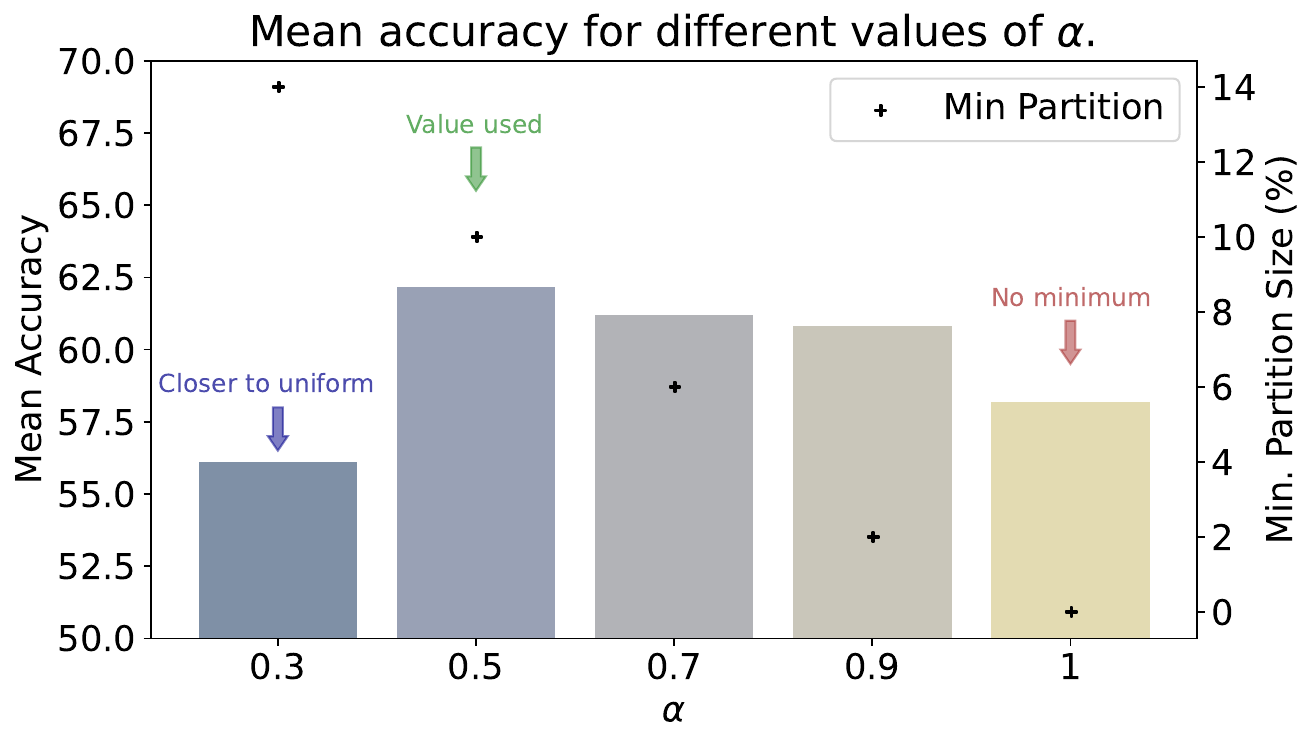}
    \caption{\textbf{Left: } Dual variables indicate the sensitivity of the performance on the current task with respect to the no-forgetting requirement enforced on past tasks (Theorem \ref{theo:omegasubdiff}). \textbf{Right: } Impact of minimum enforced partition size in speech classification (see Section \ref{sec:abp}).}
    \label{fig:alpha}
\end{figure}

Theorem \ref{theo:omegasubdiff} implies that $\boldsymbol{\lambda}^{\star}$ yields a near global linear under-estimator of $\tilde{P}_t$ at $\epsilon$. In particular, the impact of a perturbation $\gamma_k = [0,\cdots,0,\gamma,0,\dots,0]$ of the forgetting tolerances $\epsilon$ is described by
$$
\tilde{P}_t(\epsilon + \gamma_k) - \tilde{P}_t(\epsilon) \geq  -\lambda^{\star}_k \gamma - \omega. 
$$ 
This means that the dual variable $-\lambda^{\star}_k$ carries information about the relative difficulty of task $k$. Specifically, tightening the constraint associated to task $k$ ($\gamma<0$) restricts the feasible set, causing a degradation of the optimal value of \eqref{Ptilde} (i.e., the performance on the current task) at a rate larger than $\lambda^{\star}_k$, with an offset of $\omega$. In this sense, optimal dual variables reflect how hard it is to achieve good performance in the current task (\emph{plasticity}), while maintaining the performance on a previous task (\emph{stability}). Similarly, relaxing the constraint associated to a certain task $k$ (i.e., $\gamma>0$) would expand the feasible set, leading to a potential improvement of the expected loss in the current task of at most $\lambda^{\star}_k \gamma + \omega$.  In this sense, $\lambda^{\star}_k$ captures the stability-plasticity trade-off associated to task $k$. 

Theorem \ref{theo:omegasubdiff} indicates that, as the capacity of the model increases, the constant $\nu$ decreases and the quality of $\boldsymbol{\lambda}^{\star}$ as a sensitivity indicator improves. This emphasizes \emph{the importance of rich parametrizations in (constrained) learning}, which has been established by \citep{arora2018optimization, safran2021effects, elenter2024nearoptimal, bubeck2021universal}, among others. Conversely, as we reduce the size of the memory buffer, potentially decreasing $\tilde{n}$, the sample complexity function $\zeta(\tilde{n}, \delta)$ grows and degrades the information captured by $\boldsymbol{\lambda}^{\star}$. This is not unlike other CL strategies that suffer in small budget regimes \citep{chaudhry2019tiny}, particularly in settings with low task similarity.

\subsection{Adaptive Buffer Partition}
\label{sec:abp}

\begin{figure}[b]
\centering
\hspace{-0.2in}
\includegraphics[width=.30\linewidth]{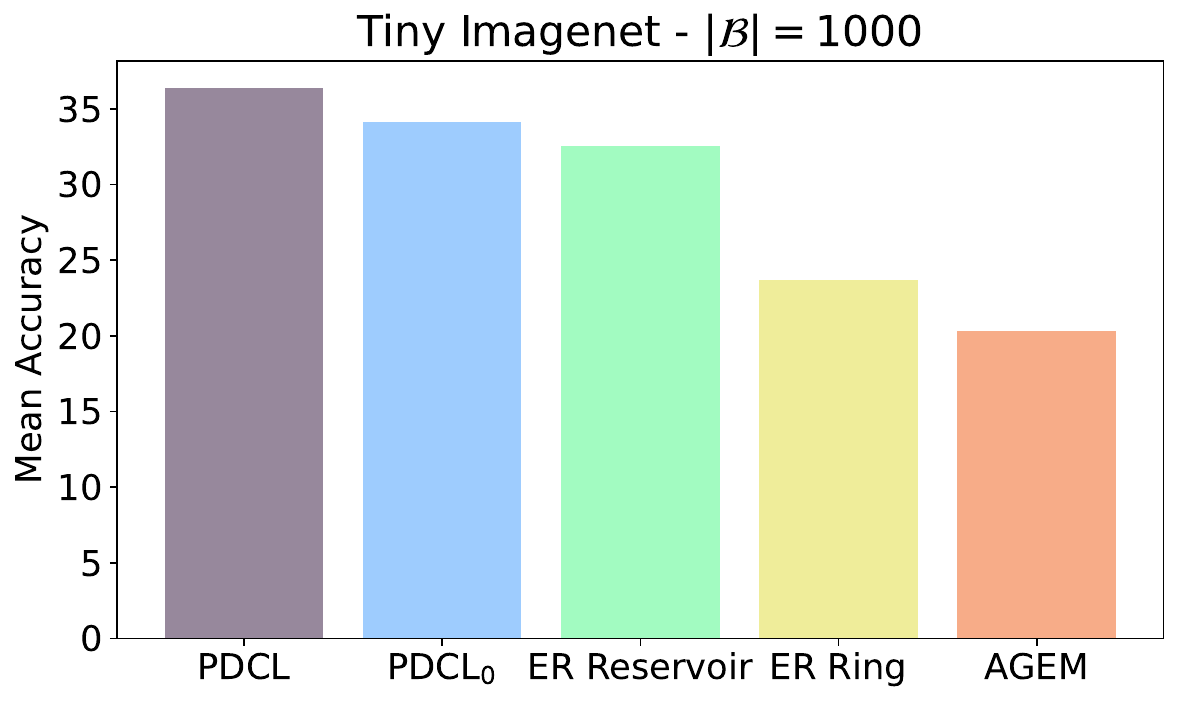}
\hspace{0.1in}
\includegraphics[width=.30\linewidth]
{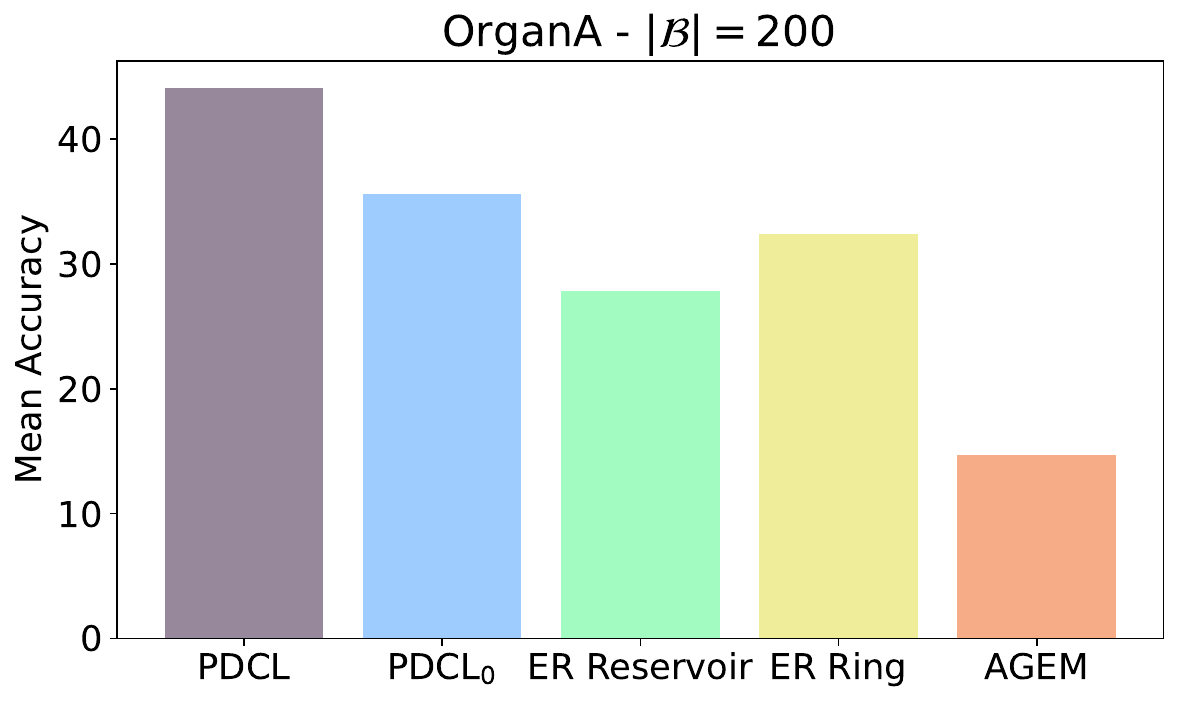}
\hspace{0.1in}
\includegraphics[width=.30\linewidth]{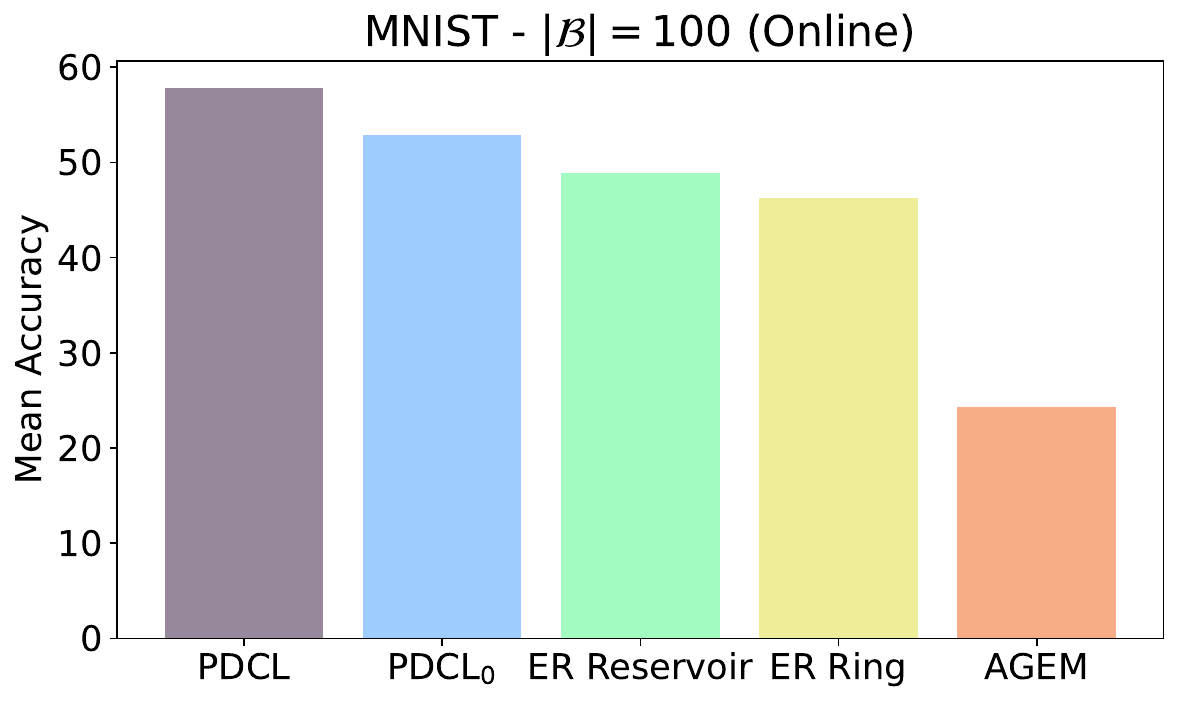}

\caption{Leveraging dual variables (PDCL$_0$) to adaptively weight the replay losses provides an improvement over fixed regularization weights (ER Ring and Reservoir). Partitioning the buffer non-uniformly (PDCL) according to the task difficulty measured by $\mathbf{\lambda}$ improves over uniform (PDCL$_0$, ER Ring, AGEM) partitions. Gradient projections (AGEM) tend to perform worse than replays.}
\label{fig:buffpart}
\end{figure} 

In light of Theorem \ref{theo:omegasubdiff}, it is sensible to \emph{partition the buffer across different tasks as an increasing function of $\boldsymbol{\lambda}^{\star}$}, allocating more resources to tasks with higher associated dual variable. That is, $  \boldsymbol{n}(t) = |\mathcal{B}| \frac{\boldsymbol{\lambda}(t)}{ \| \boldsymbol{\lambda}(t) \|_1}
$. To prevent the potential issue of allocating no samples to a task when $\boldsymbol{\lambda} = 0$ we impose a lower bound on the partition size. This contemplates the fact that tasks deemed easy at a certain iteration $t$ might eventually limit the performance of a future task $t'$. This leads to a buffer partition given by the following affine map on $\boldsymbol{\lambda}(t)$ :
\begin{equation}
\label{PB}
\tag{PB}
PB(\boldsymbol{\lambda}(t)) = |\mathcal{B}| \left( \alpha \frac{\boldsymbol{\lambda}(t)}{ \| \boldsymbol{\lambda}(t) \|_1}  + \frac{1-\alpha}{ t } \right).
\end{equation}
with $\alpha \in [0, 1]$. The larger the value of $\alpha$ the smaller the lower bound. For instance, setting $\alpha=1/2$ (value used in experiments) guarantees a minimum partition size  of $|\mathcal{B}|/2t$ samples at iteration $t$, while $\alpha=1$ gives a minimum partition size of $0$. This strategy leads to a dynamic partition of the buffer that prioritizes sensitive tasks, as measured by how much they limit the performance on the current one. We track the accumulation of the slacks associated to the current task in an artificial dual $\lambda_t(t)$ (i.e, not associated to a constraint) and concatenate it to $\boldsymbol{\lambda}(t)$ to compute the partition of the current task. A more intricate partition strategy that contemplates both the sensitivity information provided by $\boldsymbol{\lambda}$ and the generalization gap is described in Appendix \ref{app:genpartition}.

\begin{figure}[t]
\centering
\includegraphics[width=.34\linewidth]{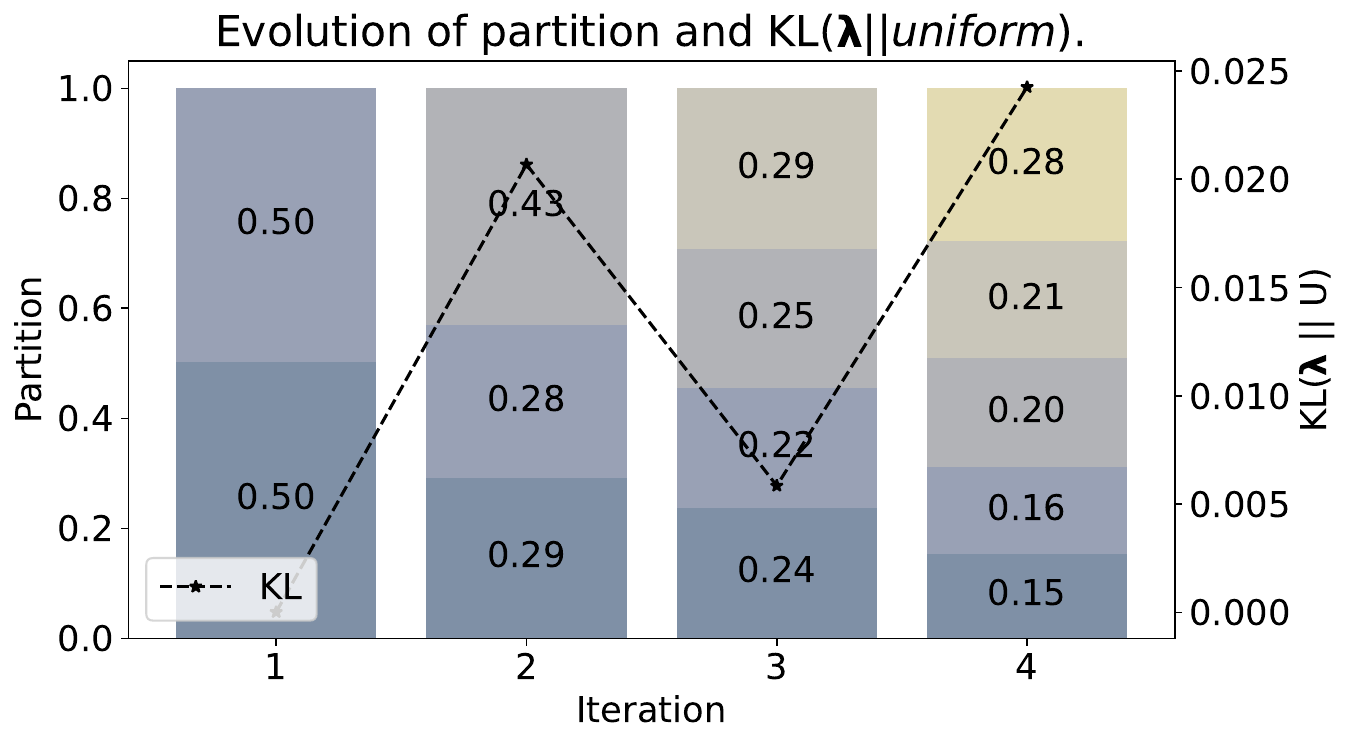}
\hfill
\includegraphics[width=.29\linewidth]{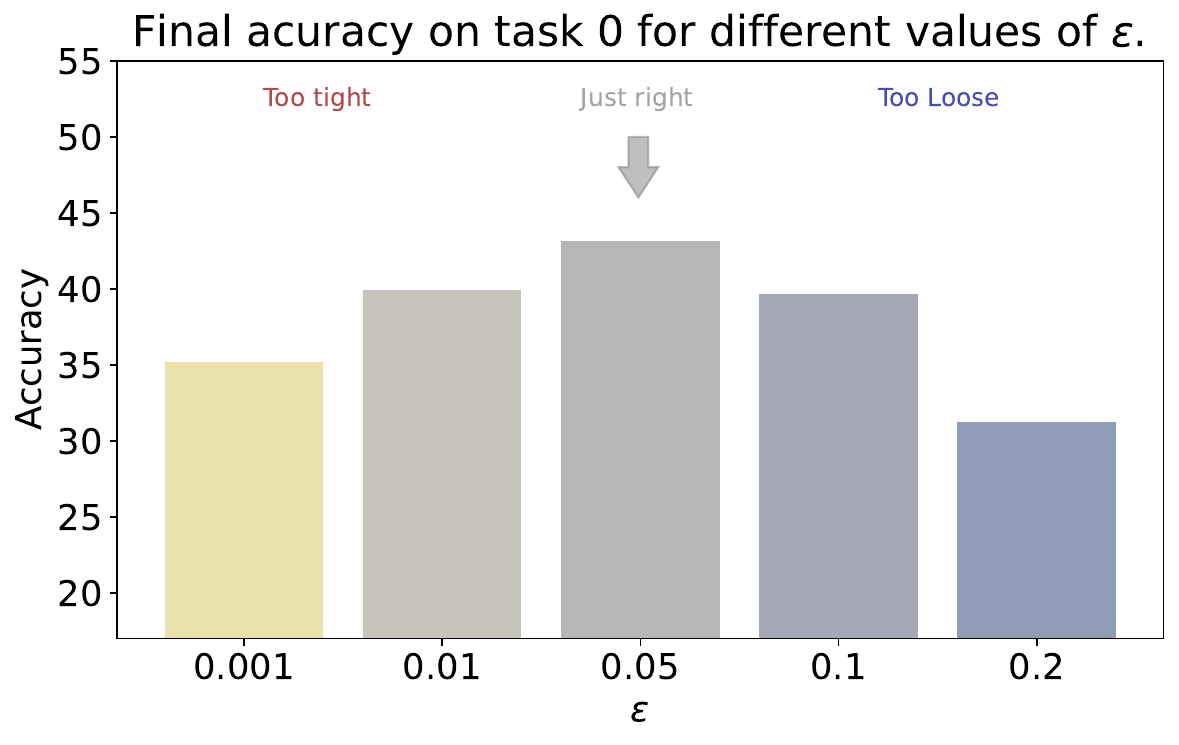}
\hfill
\includegraphics[width=.34\linewidth]{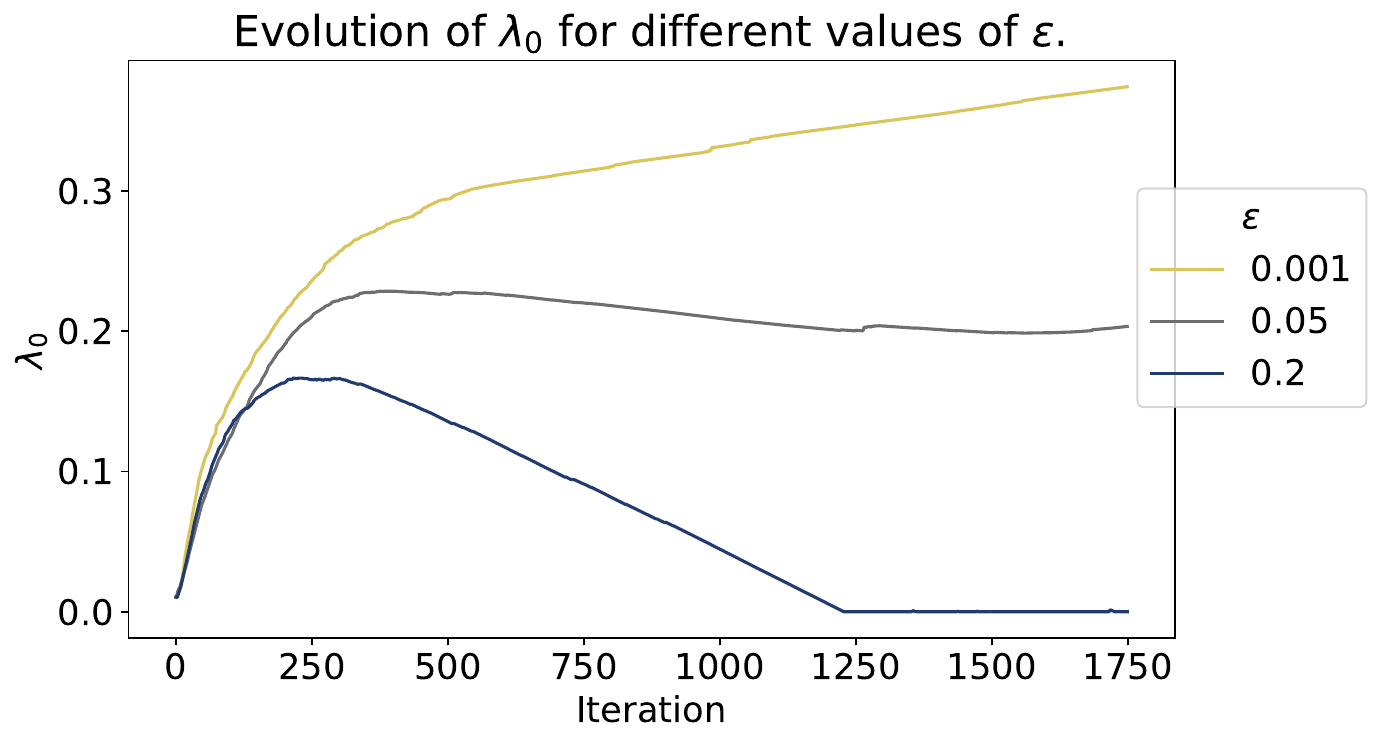}
\caption{\textbf{Left: }Evolution of non-uniform partitions obtained by (PDCL) in OrganA, and its distance from a uniform one. \textbf{Center: } If $\epsilon$ is set too loose, we allow larger forgetting. Conversely, if $\epsilon$ is too tight, (\ref{Pt}) can become harder to solve due to the reduction of its feasible set. \textbf{Right: } For a tight $\epsilon$, we violate the constraint and the associated dual variable can grow indefinitely. For a loose $\epsilon$, $\lambda_0\to0$, indicating that the performance on task 0 is not limiting learning the current task.}
\label{fig:organaparty}
\vspace{-0.1in}
\end{figure}

\section{Margin Aware Sample Selection}
\label{sec:sample_selec}

When filling the buffer by sampling uniformly at random from each observed dataset, there is no sampling bias. That is, the distributions $\mathfrak{B}_k(x, y)$ underlying the memory buffer and the data distributions $\mathfrak{D}_k(x, y)$ match. In this case, the solution of \eqref{Pt} has no-forgetting PAC learning guarantees; see e.g., \citep[Theorem 1]{peng2023ideal}. Nevertheless, inducing a bias in the buffer distribution $\mathfrak{B}_k(x, y)$ by selecting which samples to store can be beneficial due to the following:
\begin{itemize}
    \item The \emph{i.i.d.} assumption may not hold, in which case sample selection has theoretical and empirical benefits, particularly as an outlier detection mechanism \citep{sun2022informationtheoretic, peng2023ideal, coresets}. 
    \item Random sampling is not optimal in terms of expected risk decrease rate, which is the main property exploited in active and curriculum learning \citep{al_survey, gentile2022fast, elenter2022a}. In particular, \cite{hacohen2022active} suggest that easy samples are preferred in low-budget regimes, while hard instances provide better results above a certain budget threshold. 
\end{itemize}

\subsection{Identifying Impactful Samples} 
\label{ident} 
%\begin{figure}[h]
%\vspace{-0.2in}
%\centering
%\begin{subfigure}{.45\textwidth}
%  \centering
%  \includegraphics[width=.8\linewidth]{figures/selection/accuracy_degradation.pdf}
%  \caption{Accuracy degradation when storing all samples with $\lambda>0$ vs $\lambda=0$.}
%  \label{fig:sample1}
%\end{subfigure}
%\hfill
%\begin{subfigure}{.45\textwidth}
%  \centering
%  \includegraphics[width=.7\linewidth]%{figures/selection/nice_dataset_0_images_tsne.pdf}
  %\caption{t-SNE map two MNIST classes. Size of the marker %indicates associated dual variable $\lambda(x, y)$. }
  %\label{fig:sample2}
%\end{subfigure} 
%\caption{Informativeness of dual variables for sample selection. In (b), samples with large associated dual variables tend to accumulate in the task decision boundary and edges of the class cluster, while (a) shows that storing all of these samples, as opposed to storing all of those in the center of the cluster, is beneficial in terms of forgetting. }
%\end{figure}

\begin{wrapfigure}{r}{0.32\textwidth}
\vspace{-0.28in}
\centering
\adjustbox{trim=2mm 2mm 2mm 2mm, clip}{\includegraphics[width=\textwidth]{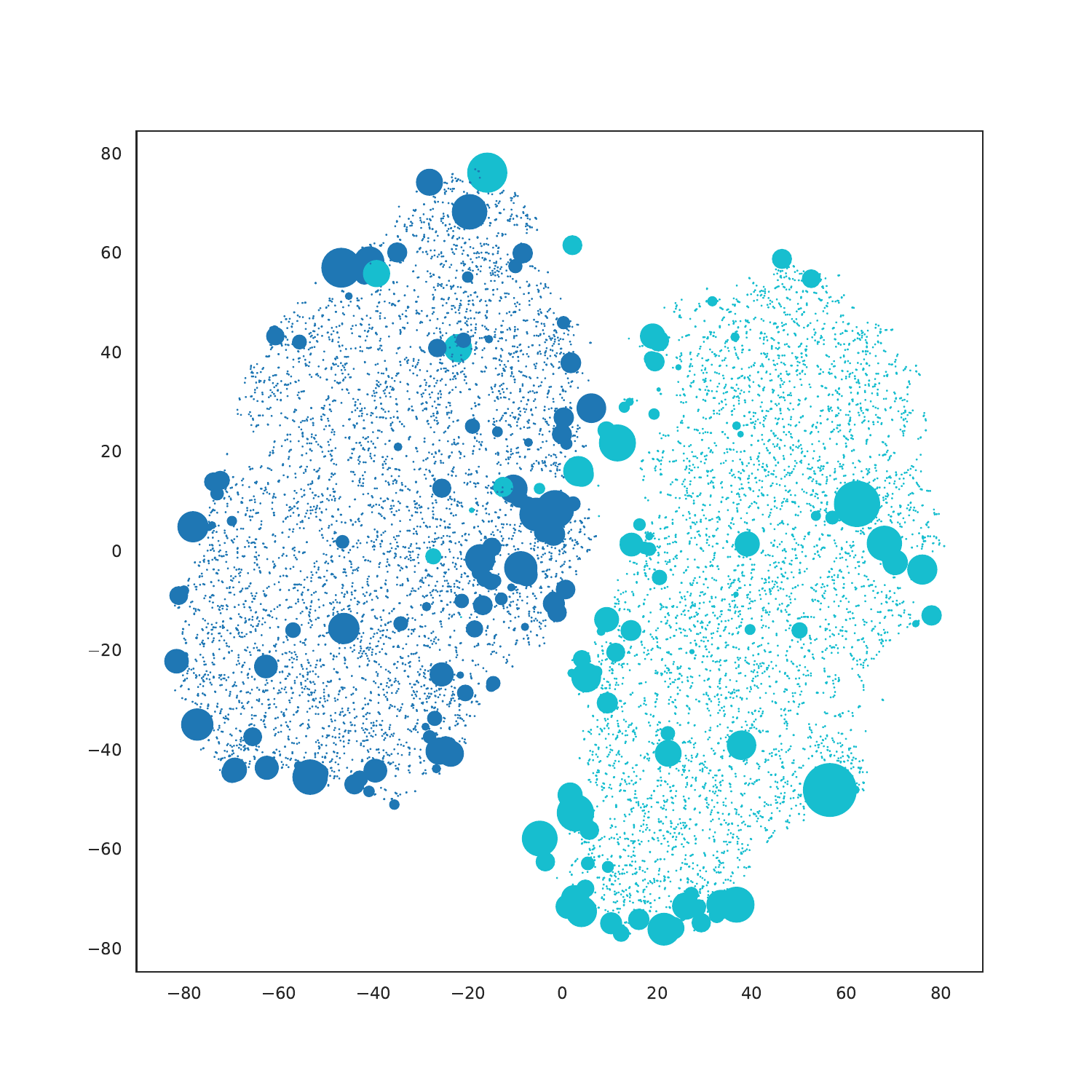}}
\caption{Class clusters with $\lambda_{x, y}$ indicated by marker size. Large dual variables accumulate in the task decision boundary and edges cluster.}
\label{fig:tsne}
\end{wrapfigure}

Instead of task-level constraints, we can formulate continual learning as enforcing a no-forgetting requirement at the sample level. For a given tightness $\epsilon$, this constraint is stricter than the task-level constraint and enables sample selection. This no-forgetting requirement can be written as:
\begin{align*}
\label{St}
\tag{$S_t$}
S_t = & \min_{\theta \in \Theta} \: \mathbb{E}_{\mathfrak{D}_t} [ \ell(f_\theta(x), y)  ], \\
& \: \text{s.t. } \:\: \ell(f_\theta(x), y) \leq \epsilon_{x, y}, \quad \mathfrak{B}_k(t)\text{-a.e.} \quad \forall \: k=1,\dots,t-1,
\end{align*}
where `a.e' (almost everywhere) means that the constraint should hold for all $(x, y)$ except possibly a set of $\mathfrak{B}_k(t)$-measure $0$. As described in the previous section, non-uniform sampling induces a bias in the buffer distribution. In what follows, we put forward a dual variable-based sampling strategy, in the same vein as the buffer partition strategy, that is beneficial in terms of expected risk reduction. We refer to \citep{farquhar2021statistical} for a general theoretical analysis of the distributional bias induced by active sampling in the context of replay methods.

The dual update rule (Line 8) in the sample-wise version of Algorithm \ref{alg:pdcl} is given by:
$$ \lambda_{x, y} \leftarrow \left[ \lambda_{x, y} + \eta_d(\ell(f_{\theta}(x), y) - \epsilon_{x, y}) \right]_+ .$$
Thus, in this formulation, dual variables accumulate the sample-wise constraint slacks over the entire learning procedure. This allows dual variables to be used as a measure of sample informativeness, while at the same time affecting the local optimum to which the primal-dual algorithm converges. Similar ideas on monitoring the evolution of the loss---or training dynamics---for specific samples in order to recognize impactful instances have been used in generalization analyses~\citep{forgetting, notallsamples} and active learning~\citep{train_dynamics, elenter2022a}. In this case, a similar sensitivity analysis as in Section \ref{sec:duals_stabplast} holds at the sample level:

\begin{corollary}
\label{coro:sample_level_sensitivity}
Let $\Tilde{S}_t$ denote the unparametrized optimal value function of problem \ref{St}. Under Assumptions \ref{ass:nu} and \ref{ass:const_qual}, with probability at least $1-t\delta$, $\boldsymbol{\lambda}^{\star}_{x, y}$ belongs to the $\omega-$subdifferential of $\Tilde{S}_t(\epsilon)$ at $\epsilon$. That is, $- \boldsymbol{\lambda}^{\star}_{x, y} \: \in \: \partial_{\omega}\Tilde{S}_t(\epsilon_{x, y})$ with $\omega(\delta)$ as in Theorem \ref{theo:omegasubdiff} .
\end{corollary}

As in the task-level constraints, Corollary~\ref{coro:sample_level_sensitivity} implies that the constraint whose perturbation has the most potential impact on $S_t$ is the constraint with the highest associated optimal dual variable. As such, infinitesimally tightening the constraint in a neighborhood $(x, y)$ would restrict the feasible set, causing an increase of the optimal value of $S_t$ at a rate larger than $\lambda_{x, y}$. In that sense, the magnitude of dual variables can be used as a measure of informativeness of a training sample. Similarly to \emph{non-support vectors in SVMs}, samples associated to inactive constraints (i.e., $\{ (x, y) : \lambda^{\star}_{x, y}(t) = 0\}$), are considered uninformative. This notion of informativeness is illustrated in Figure~\ref{fig:tsne}. 

\subsection{Dual Variable Based Sample Selection}
In Primal-Dual Continual Learning with Sample selection (PDCL-S), the buffer is filled by leveraging the per-sample dual variables $\lambda_{x, y}$, which act as an informativeness score, indicating the sensitivity of the performance on current task with respect to each stored sample. We then interpret $\boldsymbol{\lambda}(t) / \| \boldsymbol{\lambda}(t) \|_{1}$ as a probability distribution over $\mathcal{X}\times\mathcal{Y}$, where the probability of storing a pair $(x, y)$ is given by the magnitude of its dual variable $\lambda_{x, y}(t)$. Specifically, given a buffer partition $n_1, \cdots, n_t$, the generic mechanism $FB$ for populating the buffer in Algorithm~\ref{alg:pdcl} is implemented by sampling without replacement from this discrete distribution. Thus, (PDCL-S) aims to match the buffer-induced distribution $\mathfrak{B}(t)$ and the optimal dual variable function $\boldsymbol{\lambda}(t)$. 
\vspace{-0.2in}

\begin{equation}
\label{FB}
\tag{FB}
\mathcal{B}_k(t) \leftarrow FB(\boldsymbol{\lambda}(t), \boldsymbol{n}(t)) = \left\{  (x_1, y_1), \dots, (x_{n_k(t)}, y_{n_k(t)}) \right\} \sim \frac{\boldsymbol{\lambda}_k(t)}{\| \boldsymbol{\lambda}_k(t) \|_1}
\end{equation} 
where $\boldsymbol{\lambda}_k(t) = \left[ \lambda_{x, y}(t) : (x, y) \in \mathcal{B}_k(t) \right]$ denotes the vector of dual variables associated to the samples in the $k^{th}$ partition of the memory buffer. Observe that the procedure of defining informativeness scores (e.g, output entropy \citep{al_survey}, gradient norm \citep{BADGE}, expected risk reduction \citep{elenter2022a}) to perform sample selection is ubiquitous in active learning. 

\section{Experimental Validation}
\label{sec:emp_valid}

\begin{figure}[t]
\centering
\includegraphics[width=.24\linewidth]{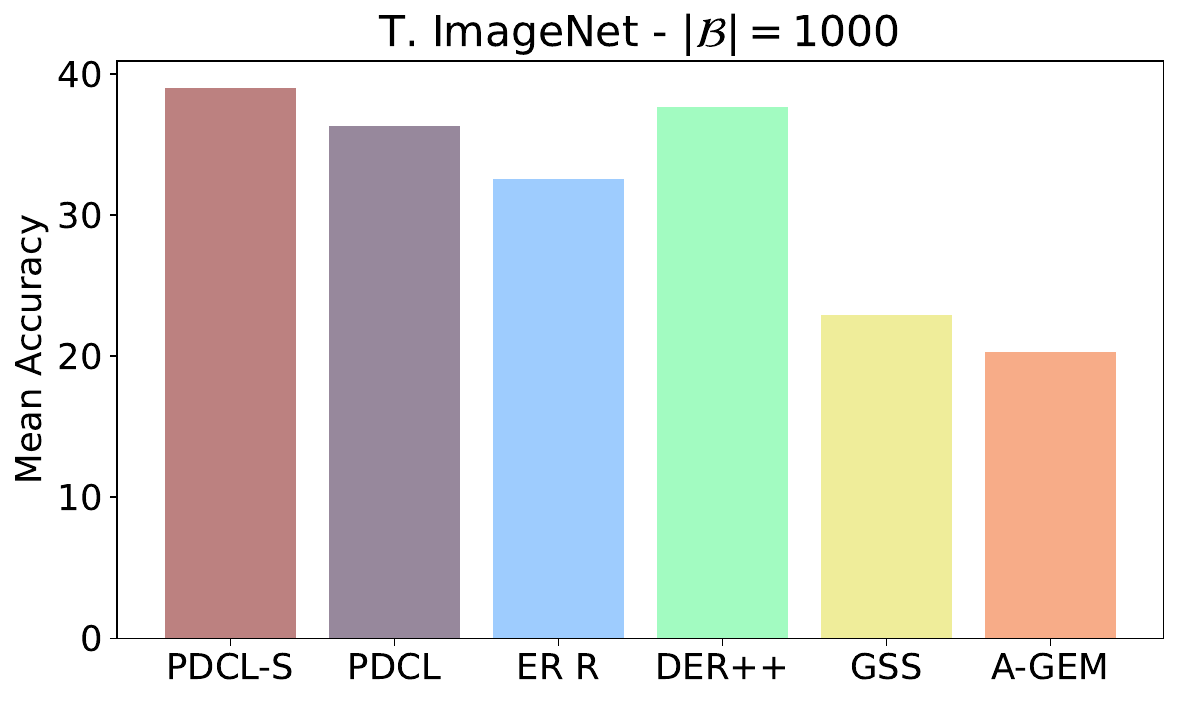}
\includegraphics[width=.24\linewidth]{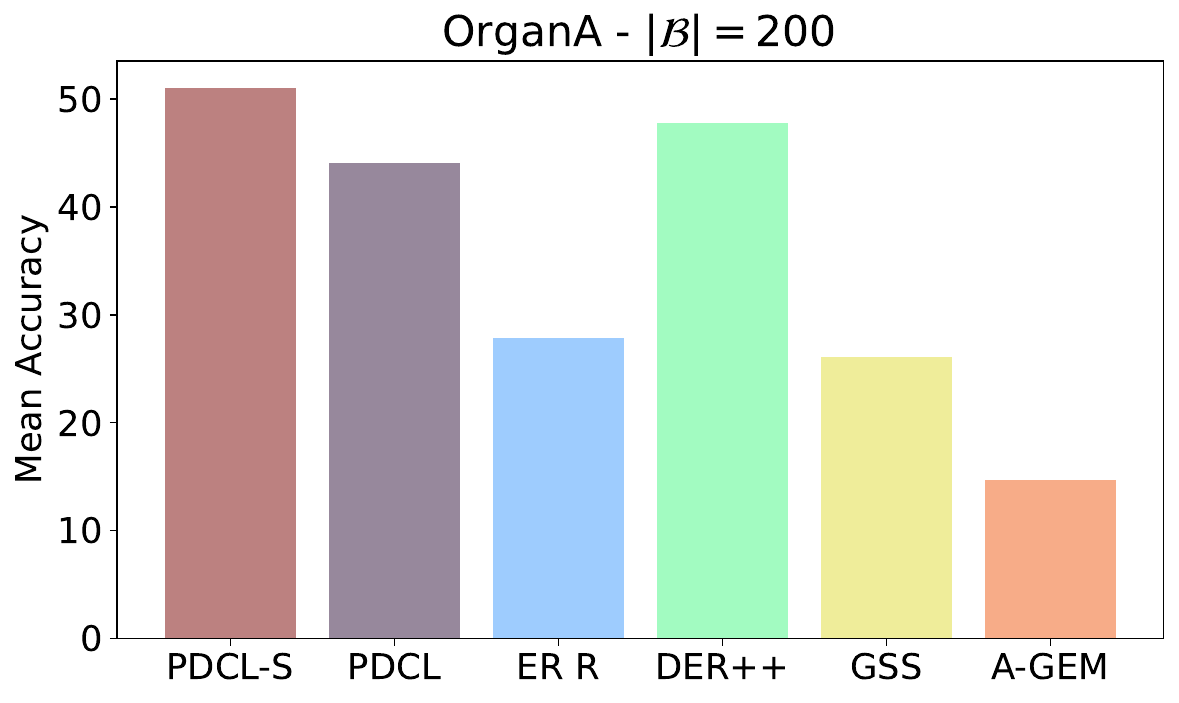}
\includegraphics[width=.24\linewidth]{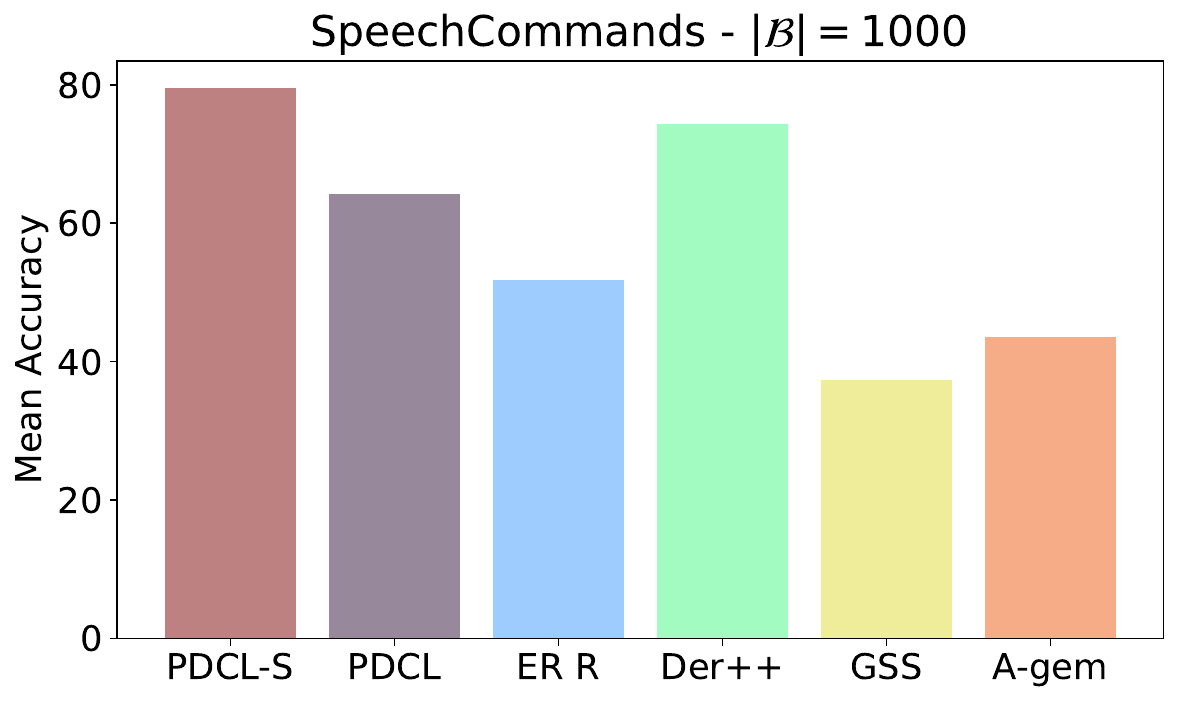}
\includegraphics[width=.24\linewidth]{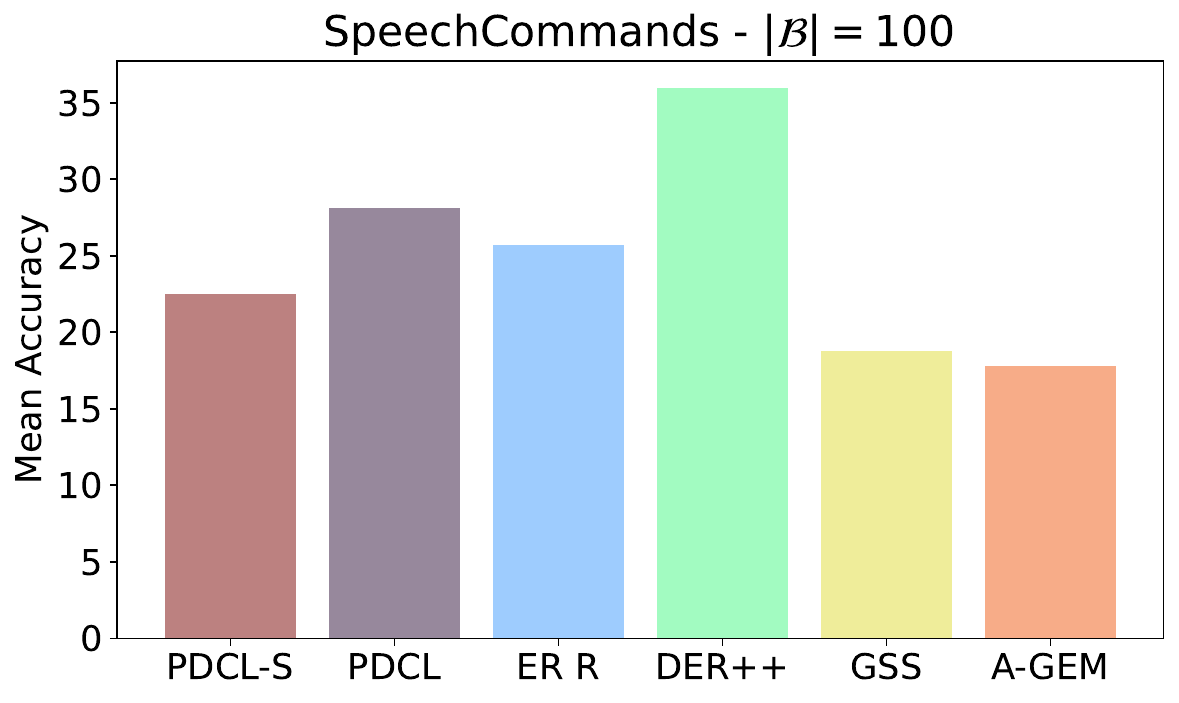}

\includegraphics[width=.24\linewidth]{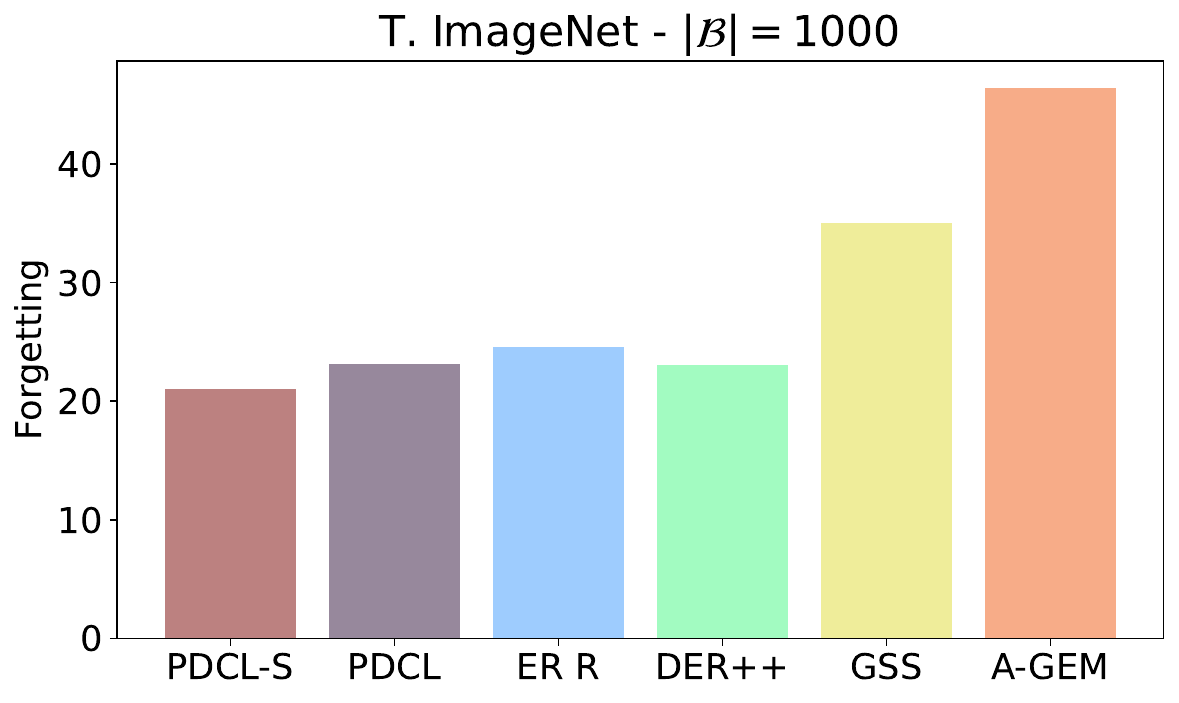}
\includegraphics[width=.24\linewidth]{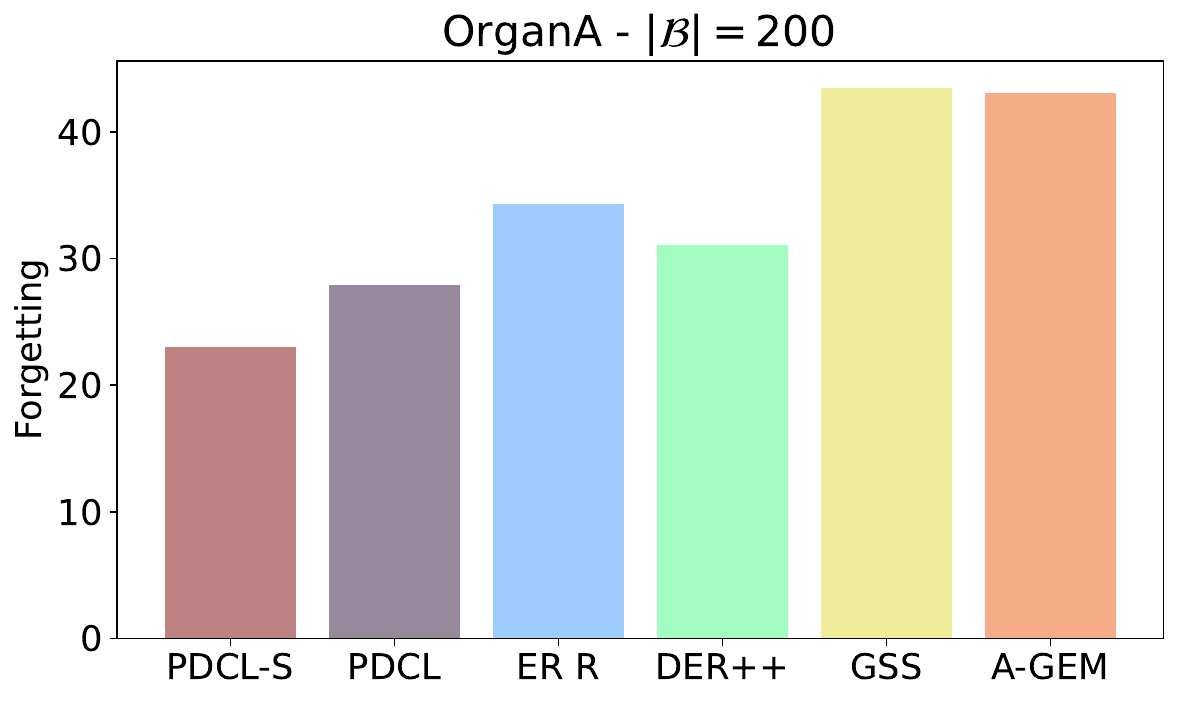}
\includegraphics[width=.24\linewidth]{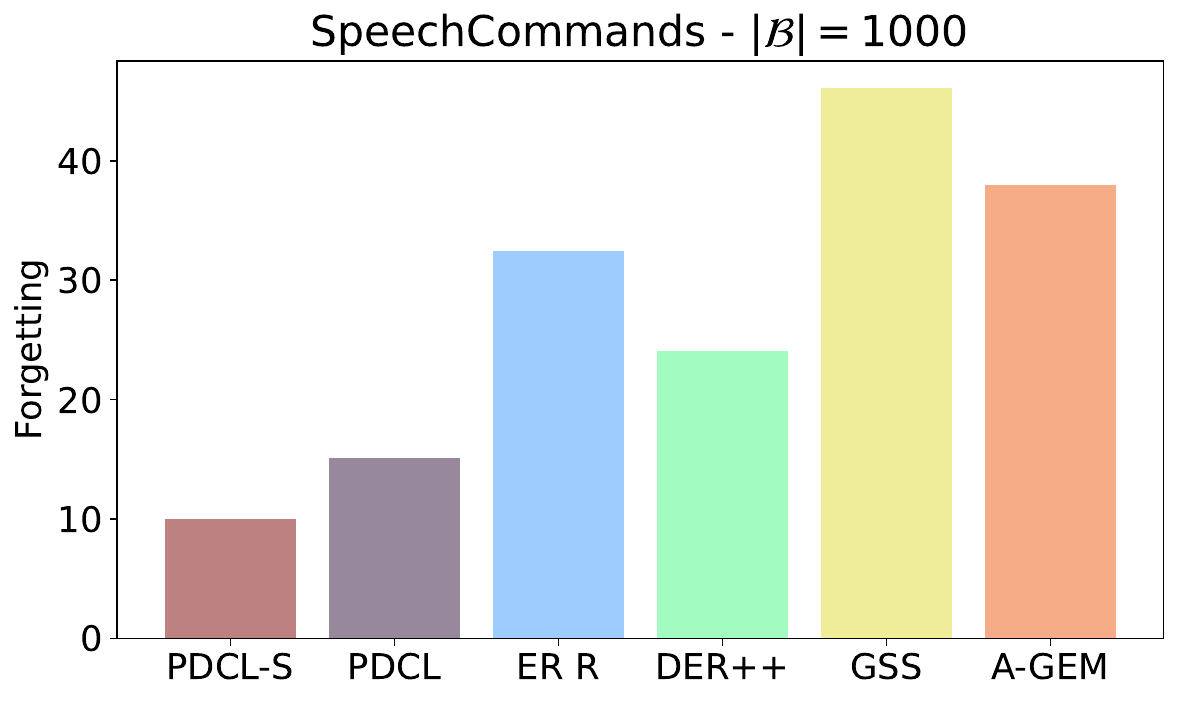}
\includegraphics[width=.24\linewidth]{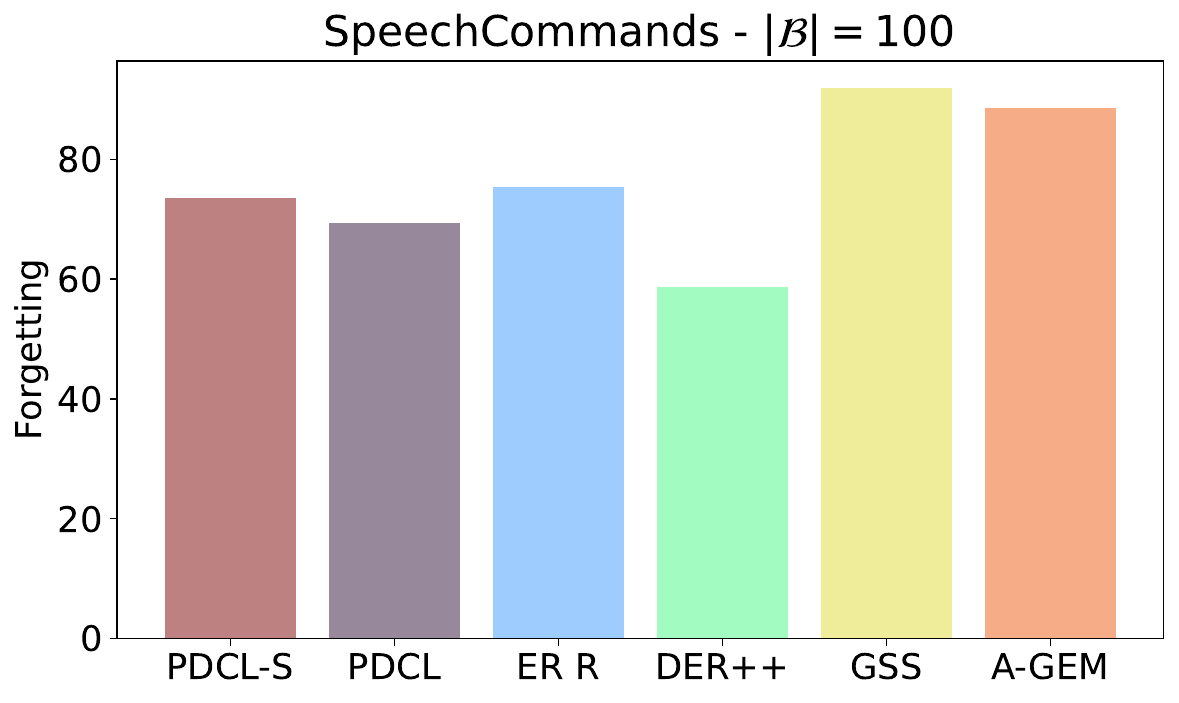}

\caption{\textbf{Top row} shows mean accuracy (higher is better) and \textbf{Bottom row} average forgetting (lower is better) across tasks after the last iteration. For a given task $i$, its \emph{Forgetting} \citep{agem} after observing task $j$ is the difference between the maximum performance level achieved in the past and the current one, i.e: $\text{Forgetting}_i(j) = \max _{l \in \{1, \cdots, j-1\}} \left( \text{acc}_i(l)-\text{acc}_i(j) \right)$. }
\label{fig:sampselec}
\end{figure} 

We evaluate the proposed strategy in four continual learning benchmarks of diverse structures: An image classification task using Tiny-ImageNet \citep{tiny_im}, a speech classification task with the SpeechCommands dataset \citep{speechcommands}, a medical image (Abdominal CT Scan)  task (OrganA \citep{organa}) and the simpler sequential MNIST \citep{mnist} dataset where we consider the online CL setting \citep{gss}. We follow the Class Incremental Learning protocol \citep{van2019three}, where datasets are split into disjoint sets, each containing a subset of the classes. MNIST, SpeechCommands, and OrganA are split into 2 class tasks, while Tiny-ImageNet is split into tasks with 4 classes. We follow \citep{buzzega2020dark} and match the model complexity to the difficulty of the problem at hand. In MNIST, we use a three-layer MLP with ReLU activations. In Seq. SpeechCommands and OrganA, we use 1-D and 2-D CNN respectively, with ReLU activations, Batch Normalization and MaxPooling. In TinyImagenet, we use a ResNet-18 architecture \citep{resnet}. At each iteration $t$, models are trained using $f_{t-1}$ as initialization. We adopt the baseline implementations of Mammoth\footnote{\url{https://github.com/aimagelab/mammoth}}, as done in \citep{darkexp}, and use the reported hyperparameters for the baselines. We report the final average accuracy and forgetting across five different random seeds. Additional experimental details are provided in Appendix \ref{app:exp_details}.

We analyze three versions of our approach: (i) PDCL$_0$: which uses dual variables as regularization weights for the replay losses, (ii) PDCL: which incorporates adaptive buffer partitioning, and (iii) PDCL-S: which also performs sample selection. As baselines, we consider the continual learning strategies presented in Appendix \ref{app:rel_work} that are most related to our work, namely, Experience Replay \citep{expreplay}, X-DER \citep{boschini2022class}, A-GEM \citep{agem}, GSS\citep{gss}. A-GEM and GSS are particularly relevant due to their similar constrained optimization perspective, and ER and X-DER are strong, standard baselines \citep{survey1}. These numerical results, showcased in Figures \ref{fig:buffpart}, \ref{fig:organaparty} and \ref{fig:sampselec}, aim to measure the impact of three factors: 
\vspace{-0.1in}
\begin{itemize}
    \item Adaptive weights (i.e., dual variables) as opposed to fixed hyper-parameters to regularize the replay losses \textbf{(ER $\to$ PDCL$_0$)}.
    \item Adaptive buffer partitioning as opposed to fixed or uniform partitions \textbf{(PDCL$_0$ $\to$ PDCL)}.
    \item Dual variable-based sample selection as opposed to uniform sampling \textbf{(PDCL $\to$ PDCL-S)}. 
\end{itemize}

We observe that undertaking the continual learning problem with a primal-dual algorithm and leveraging the information provided by dual variables leads to comparatively high mean accuracy and low forgetting in almost all buffer sizes and benchmarks. Nevertheless, in the small buffer setting of the speech processing task (right column in Figure \ref{fig:sampselec}), the performance of PDCL is modest and sample selection does not provide an improvement. This is consistent with prior work on the effectiveness of sample selection \citep{sampleselecaraujo} and with the fact that small memory buffers can negatively impact the quality of $\boldsymbol{\lambda}^{\star}$ as a sensitivity indicator (see Theorem \ref{theo:omegasubdiff}). The limited benefits of sample selection can also be attributed to the fact that, in constrained learning, problems with more constraints (sample-wise vs. task-wise) increase the condition number of the dual function, making its maximization harder \citep{elenter2024nearoptimal}. Moreover, in some settings, no method outperforms ER by a significant margin, which is consistent with recent surveys \citep{survey1}. As shown in Figure \ref{fig:tsne}, large dual variables can correspond to both outliers and inliers. Indeed, informative samples and outliers (such as mislabeled samples) may be hard to distinguish. This supports recent empirical findings \citep{outliers} indicating that many active and continual learning algorithms consistently acquire samples that most models prefer not to learn. 

%The clipping of dual variables is done beceause, as mentioned in section \ref{ident}, very high dual variables can also point to mislabelled samples or out of distribution instances. In order to avoid sampling these outliers, we discard samples in the extremely high dual variables before sampling.

\section{Conclusion}

This work demonstrates that directly addressing the constrained optimization problem in continual learning via Lagrangian duality is both feasible and beneficial. Leveraging this framework, we develop adaptive methods for managing replay buffers at both the task and sample levels, mitigating catastrophic forgetting and balancing the stability-plasticity trade-off. Our results show that dual variables capture sensitivity information about the optimization problem, enabling us to partition the buffer by allocating more resources to harder tasks and selecting the most impactful samples. These strategies lead to improved performance across diverse benchmarks. However, the quality of dual variables as sensitivity indicators degrades in small budget regimes and when the richness of the parameterization is insufficient relative to the problem's difficulty. Future work could explore several promising directions. For instance, in the context of large language models, constrained formulations of continual learning remain, for the most part, unexplored. Additionally, developing a pre-training method that yields non-uniform, feasible, and informative constraint upper bounds could significantly improve the performance of the proposed approach. Further research into the conditions under which sample selection is provably beneficial would also be valuable.

\newpage

%% file: appendix.tex
\section{Appendix}

\subsection{Related Work} % DONE
\label{app:rel_work}

Machine learning systems have become increasingly integrated into our daily lives, reaching critical applications from medical diagnostics \citep{kononenko2001machine} to autonomous driving \citep{kiran2021deep}. Consequently, the development of machine learning models that can adapt to dynamic environments and data distributions has become a pressing concern. A myriad of strategies for continual learning, also referred to as lifelong or incremental learning, have been proposed in recent years \citep{ke2023continual, guo2022online, gss, agem, ermis2022memory, icarl, expreplay, darkexp, buzzega2020dark}. In what follows, we describe some of the approaches most connected to our work. For a more extensive survey we refer to \citep{de2021continual, hadsell2020embracing}.

Two popular continual learning scenarios are task-incremental and class-incremental learning \citep{tilcil}. In task-incremental learning, the model observes a sequence of task with known task identities. These task identities have disjoint label spaces and are provided both in training and testing. This is not the case for the class-incremental setting, where task identities must be inferred by the model to make predictions. Therefore, class-incremental learning is a considerably more challenging setting \citep{masana2022class}. Continual learning methods typically fit into one of three categories: regularization-based, memory-based (also called replay methods) or architecture-based. Regularization methods \citep{ewc, si} augment the loss in order to prevent drastic changes in model parameters, consolidating previous knowledge. Moreover, architecture based methods \citep{mallya2018packnet} isolate or freeze a subset of the model parameters for each new observed task. In this work, we will focus on Memory-based methods, which store a small subset of the previously seen instances,\citep{expreplay, agem, gss} and usually outperform their memoryless counterparts \cite{survey1}. 

What mainly differentiates memory-based methods is the way in which the buffer is managed. In order to avoid forgetting, Experience Replay \citep{expreplay} modifies the training procedure by averaging the gradients of the current samples and the replayed samples. To manage the buffer, this method has two main variants: Reservoir sampling and Ring sampling. In Reservoir sampling \citep{vitter1985random}, a sample is stored with probability $B/N$, where $B$ is the memory budget and $N$ the number of samples observed so far. This is particularly useful when the input stream has unknown length, and attempts to match the distribution of the memory buffer with that of the data distribution. The Ring strategy method prioritizes uniformity among classes, and performs class-wise FIFO sampling \citep{agem}.

To select the stored instances, iCARL \cite{icarl} samples a set whose mean approximates the class mean in the feature space. During training, iCARL uses knowledge distillation and in inference, the nearest mean-of-exemplar classification strategy is performed. Some continual learning methods formulate it as a constrained optimization problem. For instance, \citep{gss} tries to find the subset of constraints that best approximate the feasible region of the original forgetting requirements. This is shown to be equivalent to a diversity strategy for sample selection. Another example is GEM \cite{gem} (or its more efficient variant AGEM \citep{agem}), where the constrained formulation leads to projecting gradients so that model updates do not interfere with the performance on past tasks. Lastly, X-DER \citep{boschini2022class} is a variant of \citep{darkexp}, usually considered a strong baseline which uses both replay and regularization strategies. X-DER promotes consistency with its past by matching the model's logits throughout the optimization trajectory. Lastly, \citep{peng2023ideal} introduced the general theoretical framework of ideal continual learners, which leads to the general constrained learning formulation.

\subsection{Sensitivity of $\tilde{P}_t$} % DONE
\label{app:sensidual}

\begin{lemma}
\label{lem:sensi}
Under Assumption \ref{ass:const_qual}, we have
\begin{equation}
  - \boldsymbol{\tilde{\lambda}}^{\star} \: \in \: \partial \tilde{P}_t(\epsilon)
\end{equation}
where $\partial \tilde{P}_t(\epsilon)$ denotes the sub-differential of $\tilde{P}_t$ with respect to $\epsilon$, and $\boldsymbol{\tilde{\lambda}}^{\star}$ is an optimal dual variable associated to problem \eqref{Dtilde}.
\end{lemma} 

\paragraph{Proof.}

We start by viewing the optimal value of problem \ref{Ptilde} as a function of the constraint tightness (or forgetting tolerance) $\epsilon_k$ associated to task $k$. Let $\epsilon = [ \epsilon_1, \dots, \epsilon_k, \dots, \epsilon_t ]$.
\begin{align*}
\label{Ptilde}
\tag{$\tilde{P}_t$}
\tilde{P}_t(\epsilon) = & \min_{\phi \in \mathcal{F}} \: \mathbb{E}_{\mathfrak{D}_t} [ \ell(\phi(x), y)  ], \\
& \, \text{s.t. } \quad \mathbb{E}_{\mathfrak{D}_k} [ \ell(\phi(x), y) ] \leq \epsilon_k, \quad \forall \: k\in\{1,\dots,t-1\},
\end{align*}
The Lagrangian $\mathcal{L}(\phi, \boldsymbol{\tilde{\lambda}}; \epsilon)$ associated to this problem can be written as 
\begin{align*}
 \mathcal{L}(\phi, \boldsymbol{\tilde{\lambda}}; & \epsilon) = \mathbb{E}_{\mathfrak{D}_t} [ \ell(\phi(x), y)  ] + \sum_{k=1}^t \tilde{\lambda}_k \left (\mathbb{E}_{\mathfrak{D}_k} [ \ell(\phi(x), y) ] - \epsilon_k \right)
\end{align*}

where the dependence on $\epsilon$ is explicitly shown. From Assumption \ref{ass:const_qual}, we have that problem \ref{Ptilde} is strongly dual (i.e: $\tilde{P}_t = \max_{\boldsymbol{\tilde{\lambda}}} \min_{\phi} \mathcal{L}(\phi, \boldsymbol{\tilde{\lambda}})$). This is because it is a functional program satisfiying Slater's constraint qualification \citep{rockafellar1997convex}. Then, following the definition of $\tilde{P}_t(\epsilon)$ and using strong duality, we have $$\tilde{P}_t(\epsilon)=\min_{\phi} \mathcal{L}(\phi, \boldsymbol{\tilde{\lambda}}^{\star}(\epsilon); \epsilon) \leq \mathcal{L}(\phi, \boldsymbol{\tilde{\lambda}}^{\star}(\epsilon); \epsilon)$$ with the inequality being true for any function $\phi \in \mathcal{F}$, and where the dependence of $\boldsymbol{\tilde{\lambda}}^{\star}$ on $\epsilon$ is also explicitly shown. Now, consider an arbitrary $\epsilon' = [ \epsilon_1, \dots, \epsilon_k', \dots, \epsilon_t ]$ which matches $\epsilon$ at all indices but $k$, and the respective primal function $\phi^{\star}(\cdot;\epsilon')$ which minimizes its corresponding Lagrangian. Plugging $\phi^{\star}(\cdot;\epsilon')$ into the above inequality, we have
\begin{align*}
    P^\star_t(\epsilon) &\leq \mathcal{L}(\phi^{\star}(\cdot;\epsilon'), \boldsymbol{\tilde{\lambda}}^{\star}(\epsilon); \epsilon) \\ &= \mathbb{E}_{\mathfrak{D}_t} [ \ell(\phi^{\star}(x; \epsilon'), y)  ] + \sum_{k=1}^t \tilde{\lambda}^{\star}_k(\epsilon) \left (\mathbb{E}_{\mathfrak{D}_k} [ \ell(\phi^{\star}(x; \epsilon'), y) ] - \epsilon_k \right)
\end{align*}
Now, since $\phi^{\star}(\cdot;\epsilon')$ is \emph{optimal} for constraint bounds given by $\epsilon'$ and complementary slackness holds, we have:
$$\mathbb{E}_{\mathfrak{D}_t} [ \ell(\phi^{\star}(x; \epsilon'), y)  ] = P^\star_t(\epsilon').$$ Moreover, $\phi^{\star}(\cdot;\epsilon')$ is, by definition, feasible for constraint bounds given by $\epsilon'$. In particular,
\begin{align*}
& \mathbb{E}_{\mathfrak{D}_k} [ \ell(\phi^{\star}(x; \epsilon'), y) ] \leq \epsilon'_k 
\end{align*}
This implies that,
\begin{align*}
 & \mathbb{E}_{\mathfrak{D}_k} [ \ell(\phi^{\star}(x; \epsilon'), y) ] - \epsilon_k \\
&= \mathbb{E}_{\mathfrak{D}_k} [ \ell(\phi^{\star}(x; \epsilon'), y) ] - \epsilon_k + \epsilon'_k - \epsilon'_k \\
&= \alpha + (\epsilon'_k - \epsilon_k) \quad \text{with } \alpha \leq 0
\end{align*}
Combining the above, we get
$$
P^\star_t(\epsilon) \leq P^\star_t(\epsilon') + \tilde{\lambda}^{\star}_k(\epsilon)( \epsilon'  - \epsilon_k ) 
$$ where we used that for all $i \neq k$, $\epsilon'_i = \epsilon_i$ and thus $\tilde{\lambda}^{\star}_i(\epsilon) \left(\mathbb{E}_{\mathfrak{D}_k} [ \ell(\phi^{\star}(x; \epsilon'), y) ] - \epsilon_i \right) \leq 0$ . Equivalently,
\begin{align*}
 P^\star_t(\epsilon')  \geq P^\star_t(\epsilon) - \tilde{\lambda}^{\star}_k(\epsilon)( \epsilon'_k  - \epsilon_k ),
\end{align*} which matches the definition of the sub-differential, completing the proof.
This result stems from a sensitivity analysis on the constraint of problem \eqref{Pt} and more general versions of it are well-known in the convex optimization literature (see e.g, \cite{shapiro_guidedtour}).

\subsection{Distance between dual functions $g(\lambda)$ and $\tilde{g}(\lambda)$}
\begin{lemma}
\label{lem:linear} The point-wise distance between the dual functions $g(\boldsymbol{\lambda}) = \min _{\theta \in \theta} \mathcal{L}(\theta, \boldsymbol{\lambda}) $ and $\tilde{g}(\boldsymbol{\lambda}) = \min _{\phi \in \mathcal{F}} \mathcal{L}(\phi, \boldsymbol{\lambda})$ is bounded as follows:
\begin{equation}
 0 \leq g(\boldsymbol{\lambda}) - \tilde{g}(\boldsymbol{\lambda}) \leq M\nu (1 + \| \boldsymbol{\lambda} \|_1 ) \quad \quad  \forall \quad \lambda \succeq 0 
\end{equation}
\end{lemma}

\paragraph{Proof. }

To alleviate the notation, we denote the statistical risks by $L_k(\phi) :=  \mathbb{E}_{\mathfrak{D}_k} [ \ell(\phi(x), y) ]$ and %the empirical risks by $\hat{L}_k(f) := \frac{1}{n_k} \sum_{i=1}^{n_k}  \ell(f(x_i), y_i)$. Similarly, 
the functional $L(\phi) := [ L_1(\phi),\dots,L_{t-1}(\phi) ]$ %and $\hat{L}(f) = [\hat{L}_1(f),\dots,\hat{L}_{t-1}(f)]$ 
collects the risks associated to past tasks. The Lagrangian can thus be written as: $$
\mathcal{L}(\phi, \boldsymbol{\lambda} ) = L_0(\phi) + \boldsymbol{\lambda}^T L(\phi).
$$

Recall that $\phi(\boldsymbol{\lambda}) = \argmin_{\phi} \mathcal{L}(\phi, \boldsymbol{\lambda} )$ denotes the Lagrangian minimizer associated to the multiplier $\boldsymbol{\lambda}$.

By the near-universality assumption \ref{ass:nu}, $\exists$ $\Tilde{\theta} \in \Theta$ such that $\| \phi(\boldsymbol{\lambda}) - f_{\Tilde{\theta}}\|_{L_2} \leq \nu$. Note that, 
\begin{align*}
    \mathcal{L}(f_{\Tilde{\theta}}, \boldsymbol{\lambda}) - \mathcal{L}(\phi(\boldsymbol{\lambda}), \boldsymbol{\lambda}) &= L_0(f_{\Tilde{\theta}}) - L_0(\phi(\boldsymbol{\lambda})) + \boldsymbol{\lambda}^T \left( L(f_{\Tilde{\theta}}) - L(\phi(\boldsymbol{\lambda}))\right) \\
    & \leq \| L_0(f_{\Tilde{\theta}}) -  L_0(\phi(\boldsymbol{\lambda})) \|_{2} + \sum_{i=1}^m [\boldsymbol{\lambda}]_i \| L(f_{\Tilde{\theta}}) - L(\phi(\boldsymbol{\lambda})) \|_{2}
\end{align*} where we used the triangle inequality twice. Then, using the $M-$Lipschitz continuity of the losses and the fact that $\| \phi(\lambda) - f_{\Tilde{\theta}}\|_2 \leq \nu$, we obtain:
\begin{align*}
    \mathcal{L}(f_{\Tilde{\theta}}, \lambda) - \mathcal{L}(\phi(\boldsymbol{\lambda}),\boldsymbol{\lambda}) &\leq M \| f_{\Tilde{\theta}} - \phi(\boldsymbol{\lambda})\|_{L_2} + M\sum_{i=1}^m [\lambda]_i \|f_{\Tilde{\theta}} - \phi(\lambda) \|_{L_2} \\
    &\leq M\nu + M\nu \sum_{i=1}^m [\lambda]_i = M\nu(1 + \|\boldsymbol{\lambda}\|_1)
\end{align*}
Since $f_{\theta}(\boldsymbol{\lambda}) \in \mathcal{F}^{\star}_{\theta}(\boldsymbol{\lambda})$ is a Lagrangian minimizer, we know that  $L(f_{\theta}(\boldsymbol{\lambda}), \boldsymbol{\lambda}) \leq L(f_{\Tilde{\theta}}, \boldsymbol{\lambda}) $. Thus,
$$ 
0 \leq \mathcal{L}(f_{\theta}(\boldsymbol{\lambda}), \boldsymbol{\lambda}) - \mathcal{L}(\phi(\boldsymbol{\lambda}), \boldsymbol{\lambda})\leq \mathcal{L}(f_{\Tilde{\theta}}, \lambda) - \mathcal{L}(\phi(\boldsymbol{\lambda}), \boldsymbol{\lambda})
$$
where the non-negativity comes from the fact that $\mathcal{F}_{\Theta} \subseteq \mathcal{F}$. This implies: 
$$ 0 \leq g(\boldsymbol{\lambda}) - \tilde{g}(\boldsymbol{\lambda}) \leq M\nu ( \| \boldsymbol{\lambda} \|_1 + 1) \quad \quad  \forall \quad \boldsymbol{\lambda} \succeq 0  $$  which conludes the proof.

\subsection{ Distance between dual iterates $\mathbf{\tilde{\lambda}}^{\star}$ and $\mathbf{\lambda}^{\star}$ }
\label{app:distduals}

The sensitivity result in Lemma \ref{lem:sensi} holds for the optimal \emph{statistical} dual variables $\boldsymbol{\tilde{\lambda}}^{\star}$ of problem \eqref{Pt}. However, in practice, we access the \emph{empirical} parameterized dual variables $\boldsymbol{\lambda}^{\star}$ of problem \eqref{Dt}. In this section, we characterize the distance between these two quantities, showing that, under mild assumptions, $\boldsymbol{\lambda}^{\star} \in \argmax \hat{g}(\boldsymbol{\lambda})$ is not far from $\boldsymbol{\tilde{\lambda}}^{\star} \in \argmax \tilde{g}(\boldsymbol{\tilde{\lambda}})$. This result depends on the curvature $\mu$ of the dual function $\tilde{g}(\tilde{\lambda})= \min_{\phi \in \mathcal{F}} \mathcal{L}(\phi, \boldsymbol{\tilde{\lambda}})$ of Problem (\ref{Ptilde}), whose characterization can be found in Appendix \ref{app:curvdual}.

\begin{proposition}
\label{prop:emp_dual}
Let $\mathcal{B}_{\lambda}$ denote the segment connecting $\boldsymbol{\tilde{\lambda}}^{\star}$ and $\boldsymbol{\lambda}^{\star}$ and let $\mu$ denote the strong concavity constant of $\tilde{g}(\lambda)$ in $\mathcal{B}_{\lambda}$. Under Assumptions \ref{ass:nu} and \ref{ass:const_qual}, with probability at least $1-t\delta$, we have:
$$
\| \boldsymbol{\lambda}^{\star} - \boldsymbol{\tilde{\lambda}}^{\star} \|_2^2 \leq \frac{2}{\mu}\left[ M\nu(1+\|\boldsymbol{\lambda}^{\star}\|_1) + 6\zeta(\Tilde{n}, \delta)(1+ \Delta) \right],
$$
where $ \Delta = \max \{ \| \boldsymbol{\tilde{\lambda}}^{\star} \|_1, \| \boldsymbol{\lambda}^{\star} \|_1 \}$, $ \Tilde{n} = \min_{i=1,\dots,t} n_i$ and the sample complexity function $\zeta\left(\tilde{n}, \delta\right)=\mathcal{O}\left( R M \sqrt{d \log \left(\Tilde{n}\right) \log (d / \delta)}^{} / {}\sqrt{\Tilde{n}} \right)$ approaches zero as $\tilde{n}$ grows.
\end{proposition}

\paragraph{Proof. }

We recall the definition of the dual function $g(\boldsymbol{\lambda}) = \min_{\theta \in \Theta} \mathcal{L}(\theta, \boldsymbol{\lambda})$ and its statistical unparametrized version $\tilde{g}(\boldsymbol{\tilde{\lambda}}) = \min_{\phi \in \mathcal{F}} \mathcal{L}(\phi, \boldsymbol{\tilde{\lambda}})$. We denote by $\hat{g}(\boldsymbol{\lambda})$ the parametrized \emph{empirical} dual function: $\min_{\theta} \hat{\mathcal{L}}(\theta, \boldsymbol{\lambda}) = \hat{L}_0(f_{\theta}) + \boldsymbol{\lambda}^T \hat{L}(f_{\theta})$.

Along with the boundedness of $\mathcal{F}$, assumption \ref{ass:const_qual} guarantees uniform convergence \citep[Theorem 5]{shalev2009stochastic}, implying that with probability at least $1-\delta$, for all $f_{\theta} \in \mathcal{F}$ we have:
$$
| L_i(f_{\theta}) - \hat{L}_i(f_{\theta}) | \leq \mathcal{O}\left(\frac{MR\sqrt{d\log(n_i)\log(|\Theta| / \delta})}{\sqrt{n_i}} \right)
 := \zeta(n_i, \delta)$$  with probability at least $1-\delta$ over a sample of size $n_i$ (see \cite{shalev2009stochastic}). 

From the $\mu-$strong concavity of $\tilde{g}(\lambda)$ in $\mathcal{B}_u$, we have that:
$$
\tilde{g}(\lambda) \leq \tilde{g}(\boldsymbol{\tilde{\lambda}}^{\star}) + \nabla \tilde{g}(\boldsymbol{\tilde{\lambda}}^{\star})^T (\lambda - \boldsymbol{\tilde{\lambda}}^{\star})  - \frac{\mu}{2} \|\lambda - \boldsymbol{\tilde{\lambda}}^{\star} \|^2  \quad \forall \lambda \in \mathcal{B}_u
$$
Evaluating at $\boldsymbol{\lambda}^{\star}$ and using that $ \nabla \tilde{g}(\boldsymbol{\tilde{\lambda}}^{\star}) = L(f(\boldsymbol{\tilde{\lambda}}^{\star}))$:
$$
\tilde{g}(\boldsymbol{\lambda}^{\star}) \leq \tilde{g}(\boldsymbol{\tilde{\lambda}}^{\star}) + L(f(\boldsymbol{\tilde{\lambda}}^{\star}))^T (\boldsymbol{\lambda}^{\star} - \boldsymbol{\tilde{\lambda}}^{\star})  - \frac{\mu}{2} \| \boldsymbol{\lambda}^{\star} - \boldsymbol{\tilde{\lambda}}^{\star} \|^2  
$$
By complementary slackness, $L(f(\boldsymbol{\tilde{\lambda}}^{\star}))^T \boldsymbol{\tilde{\lambda}}^{\star} = 0$. Then, since $f(\boldsymbol{\tilde{\lambda}}^{\star})$ is feasible and $\boldsymbol{\lambda}^{\star} \geq 0$: $L(f(\boldsymbol{\tilde{\lambda}}^{\star}))^T \boldsymbol{\lambda}^{\star} \leq 0$. Thus,
$$
\tilde{g}(\boldsymbol{\lambda}^{\star})  \leq \tilde{g}(\boldsymbol{\tilde{\lambda}}^{\star}) - \frac{\mu}{2} \| \boldsymbol{\lambda}^{\star} - \boldsymbol{\tilde{\lambda}}^{\star} \|^2  
$$
Then, using Proposition \ref{lem:linear}, we have that:
\begin{equation}
\begin{split}
     \frac{\mu}{2}\| \boldsymbol{\lambda}^{\star} - \boldsymbol{\tilde{\lambda}}^{\star} \|^2 &\leq \tilde{g}(\boldsymbol{\tilde{\lambda}}^{\star}) -  g(\boldsymbol{\lambda}^{\star}) + M\nu(1+\| \boldsymbol{\lambda}^{\star}\|_1) \\
     & = \tilde{g}(\boldsymbol{\tilde{\lambda}}^{\star}) \pm g(\boldsymbol{\lambda}^{\dagger}) - g(\boldsymbol{\lambda}^{\star}) + M\nu(1+\| \boldsymbol{\lambda}^{\star} \|_1)
    \end{split}
\end{equation} 
where $\boldsymbol{\lambda}^{\dagger} \in \argmax g(\boldsymbol{\lambda}) $. Observe that  $\tilde{g}(\boldsymbol{\tilde{\lambda}}^{\star}) -  g(\boldsymbol{\lambda}^{\dagger}) \leq 0$ since $\tilde{g}(\boldsymbol{\lambda}) - g(\boldsymbol{\lambda}) \leq 0 \quad \forall \boldsymbol{\lambda}$. Therefore, 
$$
 \frac{\mu}{2}\| \boldsymbol{\lambda}^{\star} - \boldsymbol{\tilde{\lambda}}^{\star} \|^2 \leq g(\boldsymbol{\lambda}^{\dagger}) - g(\boldsymbol{\lambda}^{\star}) + M\nu(1+\| \boldsymbol{\lambda}^{\star} \|_1)
$$
Since $\boldsymbol{\lambda}^{\star}$ maximizes its corresponding dual function, we have that $\hat{g}(\boldsymbol{\lambda}^{\dagger}) \leq \hat{g}(\boldsymbol{\lambda}^{\star})$. Then,
\begin{equation}
\label{eq:pre_empbound}
\begin{split}
     \frac{\mu}{2}\| \boldsymbol{\lambda}^{\star} - \boldsymbol{\tilde{\lambda}}^{\star} \|^2&\leq g(\boldsymbol{\lambda}^{\dagger})  \pm \hat{g}(\boldsymbol{\lambda}^{\dagger})  - g(\boldsymbol{\lambda}^{\star}) + M\nu(1+\| \boldsymbol{\lambda}^{\star} \|_1) \\
     &\leq g(\boldsymbol{\lambda}^{\dagger}) - \hat{g}(\boldsymbol{\lambda}^{\dagger}) + \hat{g}(\boldsymbol{\lambda}^{\star}) - g(\boldsymbol{\lambda}^{\star}) + M\nu(1+\| \boldsymbol{\lambda}^{\star} \|_1)
\end{split}
\end{equation}

To conclude the derivation we state the following lemma, whose proof can be found in Appendix \ref{app:proof_dist_gemp}.

\begin{lemma}
\label{lem:dist_gemp}
Under assumptions \ref{ass:const_qual}, any empirical dual function maximizer $\boldsymbol{\lambda}^{\star} \in \argmax_{\boldsymbol{\lambda} \succeq 0} \hat{g}(\boldsymbol{\lambda})$, satisfies:
\begin{equation}
    \begin{split}
    &| g(\boldsymbol{\lambda}^{\dagger}) - \hat{g}(\boldsymbol{\lambda}^{\dagger}) | \leq 3\zeta(\Tilde{n}, \delta)(1 + \| \boldsymbol{\lambda}^{\dagger} \|_1 ) \quad \quad \text{and} \\ 
    &| g(\boldsymbol{\lambda}^{\star}) - \hat{g}(\boldsymbol{\lambda}^{\star}) | \leq 3\zeta(\Tilde{n}, \delta)(1 + \| \boldsymbol{\lambda}^{\star} \|_1 ) 
    \end{split}
\end{equation}
\end{lemma}

Applying Lemma \ref{lem:dist_gemp} in equation \ref{eq:pre_empbound}, we obtain:

\begin{equation}
     \frac{\mu}{2}\| \boldsymbol{\lambda}^{\star} - \boldsymbol{\tilde{\lambda}}^{\star} \|^2 \leq  3 \zeta(\Tilde{n}, \delta)( 2 + \| \boldsymbol{\lambda}^{\star} \|_1 + \|\boldsymbol{\lambda}^{\dagger} \|_1 ) + M\nu(1+\| \boldsymbol{\lambda}^{\star} \|_1)
\end{equation} 

which concludes the proof.

%Theorem \ref{prop:emp_dual} implies that as the number of samples grows, and the capacity of the model increases (i.e., $\nu$ decreases),  $\boldsymbol{\lambda}^{\star}$ approaches $\boldsymbol{\tilde{\lambda}}^{\star}$. Thus, provided the model has enough capacity, $\boldsymbol{\lambda}^{\star}$ can be used as a sensitivity indicator of $P^{\star}_t$. Since optimal dual variables indicate the sensitivity of the optimal value with respect to constraint perturbations, the term $(1 + \Delta)$ can be seen as an indicator of the sensitivity of the optimization problem. The second term captures the effect of estimating expectations with sample means, and decreases as we include more samples in the buffer. Putting together these two results, we obtain that $\boldsymbol{\lambda}^{\star}$ belongs to the $\omega-$subdifferential of $\partial \tilde{P}_t(\epsilon)$, where the constant $\omega$ is of order $\mathcal{O}(M \nu  \|\boldsymbol{\lambda}^{\star} \|_1 )$ and captures the well-posedness of the constrained optimzation problem. 

\subsection{Proof of Theorem \ref{theo:omegasubdiff}: Information captured by dual variables}

The proof of Theorem \ref{theo:omegasubdiff} stems from a straightforward instantiation of the results in Lemma \ref{lem:sensi} and Proposition \ref{prop:emp_dual}. For completeness, we recall the statement of Theorem \ref{theo:omegasubdiff}, which shows that dual variables  $\boldsymbol{\lambda}^{\star}$ can be used as a measure of relative task difficulty.

Under Assumptions \ref{ass:nu} and \ref{ass:const_qual}, with probability at least $1-t\delta$, $\boldsymbol{\lambda}^{\star}$ belongs to the $\omega-$subdifferential of $\Tilde{P}_t(\epsilon)$ at $\epsilon$. That is:
\begin{equation*}
  - \boldsymbol{\lambda}^{\star} \: \in \: \partial_{\omega} \Tilde{P}_t(\epsilon)
\end{equation*} with the constant $\omega = \frac{2}{\mu}\left[ M\nu(1+\|\boldsymbol{\lambda}^{\star}\|_1) + 6\zeta(\Tilde{n}, \delta)(1+ \Delta) \right]$, the sensitivity parameter $ \Delta~=~\max \{ \| \boldsymbol{\tilde{\lambda}}^{\star} \|_1, \| \boldsymbol{\lambda}^{\star} \|_1 \}$, and the sample complexity given by $ \Tilde{n} = \min_{i=1,\dots,T} n_i(T)$.

\paragraph{Proof.} From Lemma \ref{lem:sensi} we have that $
- \boldsymbol{\tilde{\lambda}}^{\star} \: \in \: \partial \tilde{P}_t(\epsilon)
$ which means that for any perturbation $\gamma$ of the forgetting tolerances $\epsilon$ such that $\gamma+\epsilon \in \text{dom}(\tilde{P}_t)$, 
\begin{align}
\label{eq:theo1}
\begin{split}
    \tilde{P}_t(\epsilon+\gamma) - \tilde{P}_t(\epsilon)  &\geq  \langle - \boldsymbol{\tilde{\lambda}}^{\star}, \gamma \rangle \\
    &= - \langle \boldsymbol{\tilde{\lambda}}^{\star} \pm \boldsymbol{\lambda}^{\star}, \gamma \rangle \\
    &= - \left( \langle \boldsymbol{\tilde{\lambda}}^{\star} - \boldsymbol{\lambda}^{\star} , \gamma \rangle + \langle \boldsymbol{\lambda}^{\star}, \gamma \rangle \right)
\end{split}
\end{align}

Then, from Proposition \ref{prop:emp_dual} we know that $\boldsymbol{\lambda}^{\star} \in \argmax \hat{g}(\boldsymbol{\lambda})$ is not far from $\boldsymbol{\tilde{\lambda}}^{\star} \in \argmax \tilde{g}(\boldsymbol{\tilde{\lambda}})$. Specifically, with probability at least $1-t\delta$, we have:

\begin{equation}
\label{eq:theo2}
\| \boldsymbol{\lambda}^{\star} - \boldsymbol{\tilde{\lambda}}^{\star} \|_2^2 \leq \frac{2}{\mu}\left[ M\nu(1+\|\boldsymbol{\lambda}^{\star}\|_1) + 6\zeta(\Tilde{n}, \delta)(1+ \Delta) \right],
\end{equation}
where $ \Delta = \max \{ \| \boldsymbol{\tilde{\lambda}}^{\star} \|_1, \| \boldsymbol{\lambda}^{\star} \|_1 \}$, $ \Tilde{n} = \min_{i=1,\dots,t} n_i$ and the sample complexity function $\zeta\left(\tilde{n}, \delta\right)=\mathcal{O}\left( R M \sqrt{d \log \left(\Tilde{n}\right) \log (d / \delta)}^{} / {}\sqrt{\Tilde{n}} \right)$ 

Let $\omega^2 = \gamma \frac{2}{\mu}\left[ M\nu(1+\|\boldsymbol{\lambda}^{\star}\|_1) + 6\zeta(\Tilde{n}, \delta)(1+ \Delta) \right]$.Plugging in equation \ref{eq:theo2} in \ref{eq:theo1} and using Cauchy-Schwarz we obtain 
\begin{align}
    \begin{split}
         \tilde{P}_t(\epsilon + \gamma)  - \tilde{P}_t(\epsilon) 
&\geq   \langle \boldsymbol{\lambda}^{\star} - \boldsymbol{\tilde{\lambda}}^{\star}  , \gamma \rangle - \langle \boldsymbol{\lambda}^{\star}, \gamma \rangle \\
&\geq  - \omega |\gamma| - \langle \boldsymbol{\lambda}^{\star}, \gamma \rangle 
    \end{split}
\end{align}  with probability at least $1-t\delta$, which completes the proof.
 %We first consider the case of a constraint tightening (i.e: $\tilde{P}_t(\epsilon+\gamma) - \tilde{P}_t(\epsilon) \geq 0$), for instance, when $\gamma$ is non-positive.

\subsection{Curvature $\mu$ of the dual function $\tilde{g}(\tilde{\lambda})$}
\label{app:curvdual}

Since the dual function $\tilde{g}$ is the minimum of a family of affine functions on $\boldsymbol{\tilde{\lambda}}$, it is concave, irrespective of the non-convexity of problem \ref{Ptilde}. In constrained learning, it is standard to use weight decay (i.e, L2 regularization) in the dual domain (see e.g, \citep{hounie2023resilient}). In this case, the dual objective $\min_{\phi} L(\phi) + \boldsymbol{\lambda}^T L(\phi) - \frac{\kappa}{2} \| \boldsymbol{\lambda} \|_2^2$ becomes strongly concave, with a strong concavity constant equal to the L2 regularization parameter $\kappa$, which is typically in the order of $1.0$.

If no regularization is done in the dual domain, additional assumptions are needed to guarantee the strong concavity of the dual function. For instance, it is sufficient for the loss to be $\mu_0-$strongly convex and $\beta-$smooth and the Linear Independence Constraint Qualification (LICQ) to hold. LICQ is standard in constrained optimization and means full-rankness of the constraint Jacobian $D_{\phi} L( {\phi}(\boldsymbol{\tilde{\lambda}}^{\star}) )$ at the optimum, i.e: $\exists \sigma>0$ such that $\inf_{\| \boldsymbol{\lambda} \| = 1} \| \boldsymbol{\lambda}^T D_f L( \boldsymbol{\tilde{\lambda}}^{\star}) ) \|_2 \geq \sigma$.

Under these conditions, the dual function $\tilde{g}$ is $\mu-$strongly concave with constant
\begin{equation}
     \mu = \frac{\mu_0 \: \sigma^2 }{\beta^2(1+\Delta )^2}
\end{equation} where $\Delta~=~\max \{ \| \boldsymbol{\tilde{\lambda}}^{\star} \|_1, \| \boldsymbol{\lambda}^{\star} \|_1 \}$.

\paragraph{Proof. }

For completeness we recall two basic definitions of functional analysis used in the proof. 
\begin{definition}
We say that a functional $L_i: \mathcal{F} \to \mathbb{R}$ is \emph{Fr\'echet differentiable} at $\phi^0 \in \mathcal{F}$ if there exists an operator $D_{\phi} L_i(\phi^0) \in \mathfrak{B}(\mathcal{F}, \mathbb{R})$ such that:
$$
\lim_{h \to 0} \frac{ | L_i(\phi^0+h) - L_i (\phi^0) - \langle D_{\phi} L_i(\phi^0), h \rangle |}{ \| h \|_{L_2} } = 0
$$ 
where $\mathfrak{B}(\mathcal{F}, \mathbb{R})$ denotes the space of bounded linear operators from $\mathcal{F}$ to $\mathbb{R}$.
\end{definition}  The space $\mathfrak{B}(\mathcal{F}, \mathbb{R})$, algebraic dual of $\mathcal{F}$, is equipped with the corresponding dual norm: $$ \| B \|_{L_2} = \sup \left\{ \frac{| \langle B, \phi \rangle |} { \| \phi \|_{L_2}}  \: : \: \phi \in \mathcal{F} \, , \, \| \phi \|_{L_2} \neq 0 
 \right\} $$
which coincides with the $L_2-$norm through Riesz's Representation Theorem: there exists a unique  $g \in \mathcal{F}$ such that $B(\phi) = \langle \phi, g \rangle$ for all $\phi$  and $\|B\|_{L_2} = \|g\|_{L_2}$.

The unparametrized Lagrangian $\mathcal{L}$ has a unique minimizer $\phi(\boldsymbol{\tilde{\lambda}})$ for each $\boldsymbol{\tilde{\lambda}} \in \mathbb{R}^m_+$. Let $\boldsymbol{\tilde{\lambda}}_1, \boldsymbol{\tilde{\lambda}}_2 \in \mathcal{B}_{\lambda}$ and $\phi_1 = \phi(\boldsymbol{\tilde{\lambda}}_1), \phi_2 = \phi(\boldsymbol{\tilde{\lambda}}_2)$. 

By convexity of the functions $L_i: \mathcal{F} \to \mathbb{R}$ for $i=1,\dots,m$, we have:
\begin{align*}
    \begin{split}
    & L_i(\phi_2) \geq L_i(\phi_1) + \langle D_{\phi} L_i(\phi_1) , \phi_2 - \phi_1 \rangle , \\
    & L_i(\phi_1) \geq L_i(\phi_2) + \langle D_{\phi} L_i(\phi_2) , \phi_1 - \phi_2 \rangle
    \end{split}
\end{align*}

Multiplying the above inequalities by $[\boldsymbol{\tilde{\lambda}}_1]_i \geq 0$ and $[\boldsymbol{\tilde{\lambda}}_2]_i \geq 0$ respectively and adding them, we obtain:

\begin{equation}
\label{eq:mult_add}
    - \langle L(\phi_2) - L(\phi_1), \boldsymbol{\tilde{\lambda}}_2 - \boldsymbol{\tilde{\lambda}}_1 \rangle \geq \langle \boldsymbol{\tilde{\lambda}}_1^T D_{\phi} L(\phi_1) - \boldsymbol{\tilde{\lambda}}_2^T D_{\phi} L(\phi_2), \phi_2 - \phi_1 \rangle
\end{equation}

Since $\nabla \tilde{g}(\boldsymbol{\tilde{\lambda}}) = L(\phi(\boldsymbol{\tilde{\lambda}}))$, we have that:
\begin{equation}
\label{eq:grads}
    -\langle \nabla \tilde{g}(\boldsymbol{\tilde{\lambda}}_2)- \nabla \tilde{g}(\boldsymbol{\tilde{\lambda}}_2) , \boldsymbol{\tilde{\lambda}}_2 - \boldsymbol{\tilde{\lambda}}_1 \rangle \geq \langle \boldsymbol{\tilde{\lambda}}_1^T D_{\phi} L(\phi_1) - \boldsymbol{\tilde{\lambda}}_2^T D_{\phi} L(\phi_2), \phi_2 - \phi_1 \rangle
\end{equation}

Moreover, first order optimality conditions yield: 
\begin{align}
\label{eq:optcond}
    \begin{split}
    &D_{\phi} L_0(\phi_1) + \lambda_1^T D_{\phi}  L(\phi_1) = 0, \\
    &D_{\phi} L_0(\phi_2) + \lambda_2^T D_{\phi} L(\phi_2) = 0
    \end{split}
\end{align}
where $0$ denotes the null-opereator from $\mathcal{F}$ to $\mathbb{R}$ (see e.g: \citep{kurdila2006convex} Theorem 5.3.1).

Combining equations \ref{eq:grads} and \ref{eq:optcond} we obtain:
\begin{align}
\label{eq:strconcav}
    \begin{split}
       -\langle \nabla \tilde{g}(\boldsymbol{\tilde{\lambda}}_2)- \nabla \tilde{g}(\boldsymbol{\tilde{\lambda}}_2) ,\boldsymbol{\tilde{\lambda}}_2 -\boldsymbol{\tilde{\lambda}}_1 \rangle & \geq \langle D_{\phi} L_0(\phi_2) - D_{\phi} L_0 (\phi_1) , \phi_2 - \phi_1 \rangle \\
        & \geq \mu_0 \| \phi_2 - \phi_1 \|^2_{L_2}
        \end{split}
\end{align}
where we used the $\mu_0-$strong convexity of the operator $L_0$.

We will now obtain a lower bound on $\| \phi_2 - \phi_1 \|_{L_2}$, starting from the $\beta-$smoothness of $L_0$:
\begin{align}
\label{eq:bounds_conc1}
    \begin{split}
        \| \phi_2 - \phi_1 \|_2 &\geq \frac{1}{\beta} \| D_{\phi}L_0(\phi_2) - D_{\phi}L_0(\phi_1) \|_{L_2} \\
       & = \frac{1}{\beta} \| \lambda_2^T D_{\phi} \mathbf{\ell}(\phi_2) - \lambda_1^T D_{\phi} L(\phi_1) \|_{L_2} \\
       &= \frac{1}{\beta} \| (\boldsymbol{\tilde{\lambda}}_2 - \boldsymbol{\tilde{\lambda}}_1)^T D_{\phi} \mathbf{\ell}(\phi_2) - \boldsymbol{\tilde{\lambda}}_1^T (D_{\phi} L(\phi_1) - D_{\phi} L(\phi_2)) \|_{L_2} 
    \end{split}
\end{align}

Then, second term in the previous equality can be characterized using the LICQ assumption 
\begin{equation}
\label{eq:bounds_conc2}
     \| (\boldsymbol{\tilde{\lambda}}_2 - \boldsymbol{\tilde{\lambda}}_1)^T D_{\phi} L(\phi_2) \|_{L_2} \geq \sigma \| \boldsymbol{\tilde{\lambda}}_2 - \boldsymbol{\tilde{\lambda}}_1 \|_2 \\
\end{equation}  

For the second term, using the $\beta-$smoothness of $L_i$ we can derive:
\begin{align}
\label{eq:bounds_conc3}
\begin{split}
    \|\boldsymbol{\tilde{\lambda}}_1^T (D_{\phi} L(\phi_1) - D_{\phi} L(\phi_2)) \|_{L_2} &= \| \sum_{i=1}^m [\boldsymbol{\tilde{\lambda}}_1]_i (D_{\phi} L_i(\phi_1) - D_{\phi} L_i(\phi_2) ) \|_{L_2} \\ 
    & \leq \sum_{i=1}^m [\boldsymbol{\tilde{\lambda}}_1]_i \| D_{\phi} L_i(\phi_1) - D_{\phi} L_i(\phi_2)  \|_{L_2} \\
    & \leq \sum_{i=1}^m [\boldsymbol{\tilde{\lambda}}_1]_i \beta \| \phi_1 - \phi_2 \|_{L_2} \\
    & = \beta\|\boldsymbol{\tilde{\lambda}}_1 \|_1 \|  \phi_1 - \phi_2 \|_{L_2}
\end{split}
\end{align}
    
Then, using the reverse triangle inequality:
\begin{align}
    \begin{split}
    \| (\boldsymbol{\tilde{\lambda}}_2 - \boldsymbol{\tilde{\lambda}}_1)^T D_{\phi} L(\phi_2) - & \boldsymbol{\tilde{\lambda}}_1^T (D_{\phi} L(\phi_1) - D_{\phi} L(\phi_2)) \|_{L_2}   \\
    &\geq \| (\boldsymbol{\tilde{\lambda}}_2 - \boldsymbol{\tilde{\lambda}}_1)^T D_{\phi} L(\phi_2) \|_{L_2} - \| \lambda_1^T (D_{\phi} \mathbf{\ell}(\phi_1) - D_{\phi} L(\phi_2)) \|_{L_2} \\
    &\geq \sigma \| \boldsymbol{\tilde{\lambda}}_2 - \boldsymbol{\tilde{\lambda}}_1 \|_2 -  \beta \|\boldsymbol{\tilde{\lambda}}_1\|_1 \| \phi_2 - \phi_1 \|_{L_2} 
    \end{split}
\end{align}
Combining this with equation \ref{eq:bounds_conc1} we obtain:
\begin{align}
    \begin{split}
 & \| \phi_2 - \phi_1 \|_2 \geq \frac{1}{\beta}\left( \sigma \|\boldsymbol{\tilde{\lambda}}_2 - \boldsymbol{\tilde{\lambda}}_1 \|_2  -  \beta \|\boldsymbol{\tilde{\lambda}}_1\|_1 \| \phi_2 - \phi_1 \|_{L_2} \right) \\
&\longrightarrow \| \phi_2 - \phi_1 \|_{L_2}  \geq  \frac{\sigma}{\beta(1+\|\boldsymbol{\tilde{\lambda}}_1\|_1)} \| \boldsymbol{\tilde{\lambda}}_2 - \boldsymbol{\tilde{\lambda}}_1 \|_2 \\
    \end{split}
\end{align}
This means that we can write equation \ref{eq:strconcav} as:
$$
-\langle \nabla \tilde{g}(\boldsymbol{\tilde{\lambda}}_2)- \nabla \tilde{g}(\boldsymbol{\tilde{\lambda}}_1) , \boldsymbol{\tilde{\lambda}}_2 - \boldsymbol{\tilde{\lambda}}_1 \rangle 
 \geq \frac{\mu_0 \: \sigma^2 }{\beta^2(1+\|\boldsymbol{\tilde{\lambda}}_1\|_1)^2} \| \boldsymbol{\tilde{\lambda}}_2 -\boldsymbol{\tilde{\lambda}}_1 \|^2_2
$$

Letting $\boldsymbol{\tilde{\lambda}}_2 = \boldsymbol{\tilde{\lambda}}^{\star}$, we obtain that the strong concavity constant of $\tilde{g}$ in $\mathcal{B}_{\lambda}$ is $\mu = \frac{\mu_0 \: \sigma^2 }{\beta^2(1+\max \{  \| \boldsymbol{\tilde{\lambda}}^{\star} \|_1, \| \boldsymbol{\lambda}^{\star} \|_1 \})^2}$. A similar proof in the finite dimensional case can be found in \citep{guigues2020inexact}.

\subsection{Proof of Lemma \ref{lem:dist_gemp}}
\label{app:proof_dist_gemp}

Let $\boldsymbol{\lambda}^{\star} \in \argmax_{\lambda \succeq 0} \hat{g}(\boldsymbol{\lambda})$ be an empirical dual function maximizer. We want to show that:

\begin{equation}
    \begin{split}
    &| g(\boldsymbol{\lambda}^{\star}) - \hat{g}(\boldsymbol{\lambda}^{\star}) | \leq 3\zeta(N, \delta)(1 + \| \boldsymbol{\lambda}^{\star} \|_1 ) \quad \quad \text{and} \\ 
    &| g(\boldsymbol{\lambda}^{\dagger}) - \hat{g}(\boldsymbol{\lambda}^{\dagger}) | \leq 3\zeta(N, \delta)(1 + \| \boldsymbol{\lambda}^{\dagger} \|_1 ) 
    \end{split}
\end{equation}

Along with the boundedness of $\mathcal{F}$, assumption \ref{ass:const_qual} guarantees uniform convergence \citep[Theorem 5]{shalev2009stochastic}, implying that with probability at least $1-\delta$, for all $f_{\theta} \in \mathcal{F}$ we have:
$$
| L_i(f_{\theta}) - \hat{L}_i(f_{\theta}) | \leq \mathcal{O}\left(\frac{MR\sqrt{d\log(n_i)\log(|\Theta| / \delta})}{\sqrt{n_i}} \right)
 := \zeta(n_i, \delta)$$  with probability at least $1-\delta$ over a sample of size $n_i$ (see \cite{shalev2009stochastic}). Combining this with Holder's inequality, we obtain:
\begin{equation}
\label{eq:lagsunif}
|\langle  \boldsymbol{\lambda}, L(f_{\theta})-\hat{L}(f_{\theta}) \rangle| \leq \| \boldsymbol{\lambda}||_1 \|  L(f_{\theta})-\hat{L}(f_{\theta}) \|_{\infty} \leq \|  \boldsymbol{\lambda} \|_1 \zeta( \Tilde{n}, \delta)
\end{equation} with probability at least $1-t\delta$, where $\Tilde{n} = \min_{i=1,\cdots,t} n_i$.

For conciseness, we will denote $\hat{f_{\theta}} \in \argmin_{\theta} \hat{\mathcal{L}}(f,  \boldsymbol{\lambda})$ an empirical Lagrangian minimizer associated to the multiplier $\boldsymbol{\lambda}$ and by $f_{\theta} \in \argmin_{\theta} \mathcal{L}(f,  \boldsymbol{\lambda})$ a statistical Lagrangian minimizer associated to $ \boldsymbol{\lambda}$. Evaluating \ref{eq:lagsunif} at $(f_{\theta}(\boldsymbol{\lambda}^{\star}), \boldsymbol{\lambda}^{\star})$ and $(\hat{f_{\theta}}(\boldsymbol{\lambda}^{\star}), \boldsymbol{\lambda}^{\star})$ we obtain:
\begin{equation}
    \begin{split}
        & -\zeta(\Tilde{n}, \delta) (1+\| \boldsymbol{\lambda}^{\star} \|_1) \leq \mathcal{L}(f_{\theta}(\boldsymbol{\lambda}^{\star}), \boldsymbol{\lambda}^{\star}) - \hat{\mathcal{L}}(f_{\theta}(\boldsymbol{\lambda}^{\star}), \boldsymbol{\lambda}^{\star}) \leq \zeta(\Tilde{n}, \delta) (1+\| \boldsymbol{\lambda}^{\star}\|_1) \\
        & -\zeta(\Tilde{n}, \delta)(1+ \| \boldsymbol{\lambda}^{\star} \|_1) \leq \mathcal{L}(\hat{f}_{\theta}(\boldsymbol{\lambda}^{\star}), \boldsymbol{\lambda}^{\star}) - \hat{\mathcal{L}}(\hat{f}_{\theta}(\boldsymbol{\lambda}^{\star}), \boldsymbol{\lambda}^{\star}) \leq \zeta(\Tilde{n}, \delta) (1+ \| \boldsymbol{\lambda}^{\star} \|_1 )
    \end{split}
\end{equation}

Re-arranging and summing the previous inequalities yields:
\begin{equation}
\begin{split}
      & -2\zeta(\Tilde{n}, \delta) (1+\| \boldsymbol{\lambda}^{\star} \|_1) + \hat{\mathcal{L}}(f_{\theta}(\boldsymbol{\lambda}^{\star}), \boldsymbol{\lambda}^{\star}) - \mathcal{L}(\hat{f}_{\theta}(\boldsymbol{\lambda}^{\star}), \boldsymbol{\lambda}^{\star}) \leq g(\boldsymbol{\lambda}^{\star}) - \hat{g}(\boldsymbol{\lambda}^{\star}) \\   
      &2\zeta(\Tilde{n}, \delta) (1+\| \boldsymbol{\lambda}^{\star} \|_1) + \hat{\mathcal{L}}(f_{\theta}(\boldsymbol{\lambda}^{\star}), \boldsymbol{\lambda}^{\star}) - \mathcal{L}(\hat{f}_{\theta}(\boldsymbol{\lambda}^{\star}), \boldsymbol{\lambda}^{\star}) \geq g(\boldsymbol{\lambda}^{\star}) - \hat{g}(\boldsymbol{\lambda}^{\star})
\end{split}
\end{equation}

Using that $f_{\theta}(\boldsymbol{\lambda}^{\star})$ and $\hat{f}_{\theta}(\boldsymbol{\lambda}^{\star})$ minimize the statistical and empirical Lagrangians respectively, we can write: $\hat{\mathcal{L}}(f_{\theta}(\boldsymbol{\lambda}^{\star}),  \boldsymbol{\lambda}^{\star}) \geq  \hat{\mathcal{L}}(\hat{f}_{\theta}(\boldsymbol{\lambda}^{\star}), \boldsymbol{\lambda}^{\star})$ and $\mathcal{L}(\hat{f}_{\theta}(\boldsymbol{\lambda}^{\star}), \boldsymbol{\lambda}^{\star}) \geq \mathcal{L}(f_{\theta}(\boldsymbol{\lambda}^{\star}), \boldsymbol{\lambda}^{\star}) $. Which implies:
\begin{equation}
\begin{split}
      & -2\zeta(\Tilde{n}, \delta) (1+\| \boldsymbol{\lambda}^{\star} \|_1) +  \hat{\mathcal{L}}(\hat{f}_{\theta}(\boldsymbol{\lambda}^{\star}), \boldsymbol{\lambda}^{\star}) - \mathcal{L}(\hat{f}_{\theta}(\boldsymbol{\lambda}^{\star}), \boldsymbol{\lambda}^{\star}) \leq g(\boldsymbol{\lambda}^{\star}) - \hat{g}(\boldsymbol{\lambda}^{\star}) \\   
      &2\zeta(\Tilde{n}, \delta) (1+\| \boldsymbol{\lambda}^{\star} \|_1) + \hat{\mathcal{L}}(f_{\theta}(\boldsymbol{\lambda}^{\star}), \boldsymbol{\lambda}^{\star}) - \mathcal{L}(f_{\theta}(\boldsymbol{\lambda}^{\star}), \boldsymbol{\lambda}^{\star}) \geq g(\boldsymbol{\lambda}^{\star}) - \hat{g}(\boldsymbol{\lambda}^{\star})
\end{split}
\end{equation}

Then, using \ref{eq:lagsunif} we obtain:
\begin{equation}
      -3\zeta(\Tilde{n}, \delta) (1+\| \boldsymbol{\lambda}^{\star} \|_1)  \leq g(\boldsymbol{\lambda}^{\star}) - \hat{g}(\boldsymbol{\lambda}^{\star}) \leq 3\zeta(\Tilde{n}, \delta) (1+\| \boldsymbol{\lambda}^{\star} \|_1)  
\end{equation}

The same steps applied to $(f_{\theta}(\boldsymbol{\lambda}^{\dagger}), \boldsymbol{\lambda}^{\dagger})$ and $(\hat{f_{\theta}}(\boldsymbol{\lambda}^{\dagger}), \boldsymbol{\lambda}^{\dagger})$ yield:
\begin{equation}
    | g(\boldsymbol{\lambda}^{\dagger}) - \hat{g}(\boldsymbol{\lambda}^{\dagger}) | \leq 3\zeta(\Tilde{n}, \delta) (1+\| \boldsymbol{\lambda}^{\dagger} \|_1)  
\end{equation} which concludes the proof.

\subsection{Corollary \ref{coro:sample_level_sensitivity}}

Corollary \ref{coro:sample_level_sensitivity} is an instation of Theorem \ref{theo:omegasubdiff} for a problem with a constraint per-sample, as opposed to one constraint per task. The proof folllows the same steps as in Theorem \ref{theo:omegasubdiff}, accounting for the dimensionality of the associated dual variables. We refer to the proof in \citep[Theorem 3.2]{elenter2022a}.

\subsection{Proof of Proposition \ref{prop:set_eps}}
\label{app:prop_seteps}
%The task similarity assumption \ref{ass:task_sim} implies that for any pair of tasks $i, j$, there exists a pair of functions $f_{ij} \in \mathcal{F}_i^{\star}$ and $f_{ji} \in \mathcal{F}_j^{\star}$ such that $d(f_{ij}, f_{ji}) \leq \delta$.

Consider the average predictor $\bar{f}(x) := \frac{1}{T} \sum_{i=1}^T f_{i}(x)$.
Let $m_k$ be the uncosntrained minimum associated to a given task $k$. Then, we can write the expected loss of $\bar{f}$ as:

\begin{align}
 \mathbb{E}_{\mathfrak{D}_k} [ \ell(\bar{f}(x), y)  ]  &= \mathbb{E}_{\mathfrak{D}_k} [ \ell(\bar{f}(x), y)  ] \pm m_k \\
&=  m_k + \mathbb{E}_{\mathfrak{D}_k} [ \ell(\bar{f}(x), y)  ] - \mathbb{E}_{\mathfrak{D}_k} [ \ell(f^{\star}_k(x), y)  ] \\
&\leq  m_k + \mathbb{E}_{\mathfrak{D}_k} [ M d( \bar{f}(x), f^{\star}_k(x) )  ]  \\
&\leq  m_k + \mathbb{E}_{\mathfrak{D}_k} \left[ \frac{M}{T}  \sum_{i=1}^T d( f_{i}(x) , f^{\star}_k(x) )  \right]  
\end{align}

Then, using that $f_{i} \in \mathcal{F}^{\star}_i$ and $f^{\star}_{k} \in \mathcal{F}^{\star}_k$, and that the Haussdorf distance between these two sets is bounded by $\delta$, we can write:

\begin{equation}
 \mathbb{E}_{\mathfrak{D}_k} [ \ell(\bar{f}(x), y)  ] 
\leq  m_k + \frac{T-1}{T}M\delta \\
\end{equation}

where we used that for $i=k$, we can set $f_i = f^{\star}_k \in \mathcal{F}^{\star}_k$ and thus, that term does not contribute to the loss.

\subsection{Additional Experimental Details}
\label{app:exp_details}

At iteration $t$, models are initialized using $f_{t-1}$, except for iteration $0$, where the default PyTorch initialization for each layer (e.g, Xavier in 1D Convolutional Layers) is used. We adopt the baseline implementations of Mammoth\footnote{\url{https://github.com/aimagelab/mammoth}}, and use the reported hyperparameters for the baselines. All models are trained with Adam \citep{kingma2014adam} and, when available, we use the tuned learning rates reported.

\begin{table}[h]
    \centering
    \begin{tabular}{c|c|c|cc}
         &  $\eta_p$ & $\eta_d$ & \\ \hline
        Tiny-ImageNet & 0.01 &  0.05 & \\
        SpeechCommands & 0.001 & 0.01  & \\
        OrganA & 0.005 & 0.01 & \\
        MNIST & 0.01 & 0.1 & \\
    \end{tabular} 
    \label{tab:my_label}
\end{table}
 Dual learning rates $\eta_d$ are set at about an order of magnitude larger than their primal counterparts, which is standard in other constrained learning setups \citep{elenter2022a}. More complex dual learning schedules such as decaying step sizes (e.g, summable but not square summable) can be explored in future works. 

In SpeechCommands, audio signals are resampled to 8 kHz. We use the PyTorch implementation of 1D convolutions with kernel sizes: 80 (and a stride of 16), 3, 3, 3. The original Speech Commands dataset has 35 categories, out of which 10 are used in the reduced version. In Tiny-Imagenet, we follow the experimental setup of Mammoth and normalize the images, use Random Cropping and Random Horizontal Flips as an augmentation pipleine. We also limit the number of tasks to 10, with 4 classes each. We do not use data augmentation in SpeechCommands, OrganA or MNIST.

The forgetting tolerances $\vepsilon$ correspond to the worst average loss that one requires of a past task. In many cases, this is a \emph{design} requirement, and not a tunable parameter. For large values of epsilon, constraint slacks become negative and dual variables quickly go to zero (analogous to an unconstrained problem), which makes them uniformative. On the other hand, extremely low values of $\vepsilon$ might also be inadequate, since the tightness of these constraints can make the problem infeasible and make dual variables diverge (see Figure \ref{fig:organaparty}). The proposed method is not overly sensitive to this parameter $\vepsilon$ (see Figure \ref{fig:organaparty}) and our experiments suggest that values close to $1.1 m_k$, where $m_k$ is the average loss observed when training the model without constraints, work well in practice. By performing a grid search we select 0.08, 0.02, 0.05 and 0.005 in Tiny ImageNet, SpeechCommands, OrganA and MNIST respectively. The per-sample upper bound $\vepsilon_{x, y}$, which enforces an stricter requirement than the task level constraint, is set 10 \% larger than $\vepsilon_k$. The analysis of more complex heuristics, such as unsupervised pre-training or the use of slack variables to adjust $\epsilon_{x, y}$ is a subject of future work. The minimum buffer partition parameter $\alpha$ is set to $0.5$ (see ablation in Figure \ref{fig:alpha}).

\subsection{Generalization Aware Buffer Partition}
\label{app:genpartition}

To alleviate the notation, we denote the statistical risks by $L_k(\phi) :=  \mathbb{E}_{\mathfrak{D}_k} [ \ell(\phi(x), y) ]$ and the empirical risks by $\hat{L}_k(f) := \frac{1}{n_k} \sum_{i=1}^{n_k}  \ell(f(x_i), y_i)$. Similarly, 
the functional $L(\phi) := [ L_1(\phi),\dots,L_{t-1}(\phi) ]$ and $\hat{L}(f) = [\hat{L}_1(f),\dots,\hat{L}_{t-1}(f)]$ 
collects the risks associated to past tasks. Along with the boundedness of $\mathcal{F}$, assumption \ref{ass:const_qual} guarantees uniform convergence \citep[Theorem 5]{shalev2009stochastic}, implying that with probability at least $1-\delta$, for all $f_{\theta} \in \mathcal{F}$ we have:
$$
| L_i(f_{\theta}) - \hat{L}_i(f_{\theta}) | \leq \mathcal{O}\left(\frac{MR\sqrt{d\log(n_i)\log(|\Theta| / \delta})}{\sqrt{n_i}} \right)
 := \zeta(n_i, \delta)$$  with probability at least $1-\delta$ over a sample of size $n_i$ (see \cite{shalev2009stochastic}). 
Applying this bound, the generalization gap associated with the Lagrangian can be written as:
\begin{equation}
\left| \sum_{k=1}^{t} \lambda_k  \mathbb{E}_{\mathfrak{D}_k} [ \ell(f(x), y) ] - \sum_{k=1}^{t}  \frac{\lambda_k}{n_k}\sum_{i=1}^{n_k}  \ell(f(x_i), y_i)  \right| \leq \sum_{k=1}^{t} \lambda_k \zeta(n_k).
\end{equation} where for task $t$, we replace $\lambda_t \leftarrow 1$, so as to express the Lagrangian as a single sum.
Therefore, we can find the buffer partition that minimizes an upper bound on the Lagrangian generalization gap by solving the following non-linear constrained optimization problem,
\begin{align*}
\label{partition}
\tag{BP}
n_1^{\star}, \dots, n_t^{\star} &= \argmin_{n_1, \dots, n_t } \quad \sum_{k=1}^{t} \lambda_k  \: \zeta(n_k),  \\ & \quad \quad ~~~ \text{ s.t. } \quad \: \sum_{k=1}^{t} n_k = | \mathcal{B} |. 
\end{align*} where $\zeta\left(n_k\right)=O\left(\frac{R M \sqrt{d \log \left(n_k\right) \log (d / \delta)}}{\sqrt{n_k}}\right)$.

Since the difficulty of a task can be assessed through its corresponding dual variable, in this problem we minimize the sum of task sample complexities, weighting each one by its corresponding dual variable and restricting the total number of samples to the memory budget. 

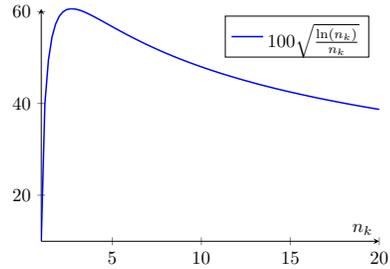
\begin{wrapfigure}{r}{0.36\textwidth}
\centering
\begin{tikzpicture}[scale=0.7]
  \begin{axis}[
    xlabel={$n_k$},    % Change the x-label to n_k and use {}
    domain=0:20,    % Define the x-axis range
    samples=100,       % Number of samples for a smooth curve
    axis lines=middle,
    legend pos=north east,
    width=8cm,         % Set the width of the figure
    height=6cm,        % Set the height of the figure
    ylabel={},         % Remove the y-label
    ]
    \addplot[blue,thick] {100*sqrt(ln(x)/x)}; % Define the function
    \legend{$100\sqrt{\frac{\ln(n_k)}{n_k}}$}
  \end{axis}
\end{tikzpicture}
\caption{Aspect of $\zeta$ function.}
\label{fig:logn}
\end{wrapfigure}

As seen in the curvature of $\zeta$ (illustrated in Figure \ref{fig:logn}), if $n_{\text{min}} = 0$ problem BP has a trivial, undesirable solution. This solution is allocating all samples to the task with highest dual variable and setting all other $n_k$ to 0. This is easily fixable, by imposing that buffer partitions should be greater than $n_\text{min}$, which corresponds to number of samples $n_k$  where $\sqrt{\frac{\log \left(n_k\right)}{n_k}}$ starts decreasing. Since $\zeta$ describes a limiting behaviour, setting a lower bound on its input is reasonable. Removing multiplicative constants that do not change the optimal argument, the  buffer partition problem can be equivalently written as:
\begin{align*}
\label{partition}
\tag{BP}
n_1^{\star}, \dots, n_t^{\star} &= \argmin_{n_1, \dots, n_t \geq n_{\text{min}}} \quad \sum_{k=1}^{t} \lambda_k \sqrt{\frac{\log \left(n_k\right)}{n_k}},  \\ & \quad \quad ~~~ \quad \text{ s.t. } \quad   \: \sum_{k=1}^{t} n_k = | \mathcal{B} | . 
\end{align*} 

In this problem, the objective adaptively weights each sample complexity $\zeta(n_k)$ by the relative difficulty $\lambda^*_k$ of task $k$. To solve this version of (BP), we can use Sequential Quadratic Programming (SQP) \cite{boggs1995sequential}, as implemented in \cite{implementslsqp}. The main idea in SQP is to approximate the original problem by a Quadratic Programming Subproblem at each iteration, in a similar vein to quasi-Newton methods. An extensive description of this algorithm can be found in \citep{gill2011sequential}